\documentclass[10pt,twocolumn,letterpaper]{article}

\usepackage{iccv}
\usepackage{times}
\usepackage{epsfig}
\usepackage{graphicx}
\usepackage{amsmath}
\usepackage{amssymb}
\usepackage{mycommands}
\usepackage{color}
\usepackage{tabularx}

\usepackage[pagebackref=true,breaklinks=true,letterpaper=true,colorlinks,bookmarks=false]{hyperref}

\newcolumntype{L}[1]{>{\raggedright\arraybackslash}p{#1}}
\newcolumntype{C}[1]{>{\centering\arraybackslash}p{#1}}
\newcolumntype{R}[1]{>{\raggedleft\arraybackslash}p{#1}}

\iccvfinalcopy 

\def\iccvPaperID{1433} 

\ificcvfinal\fi
\begin{document}

\title{Large-Scale Image Retrieval with Attentive Deep Local Features}

\author{Hyeonwoo Noh$^\dagger$ \and Andr\'{e} Araujo$^\ast$ \and Jack Sim$^\ast$ \and Tobias Weyand$^\ast$ \and Bohyung Han$^\dagger$ \and 
$^\dagger$POSTECH, Korea\\
{\tt\small \{shgusdngogo,bhhan\}@postech.ac.kr}
\and
$^\ast$Google Inc.\\
{\tt\small \{andrearaujo,jacksim,weyand\}@google.com}
}

\maketitle
\thispagestyle{empty}


\begin{abstract}

We propose an attentive local feature descriptor suitable for large-scale image retrieval, referred to as DELF (DEep Local Feature).
The new feature is based on convolutional neural networks, which are trained only with image-level annotations on a landmark image dataset.
To identify semantically useful local features for image retrieval, we also propose an attention mechanism for keypoint selection, which shares most network layers with the descriptor.
This framework can be used for image retrieval as a drop-in replacement for other keypoint detectors and descriptors, enabling more accurate feature matching and geometric verification.
Our system produces reliable confidence scores to reject false positives---in particular, it is robust against queries that have no correct match in the database.
To evaluate the proposed descriptor, we introduce a new large-scale dataset, referred to as Google-Landmarks dataset, which involves challenges in both database and query such as background clutter, partial occlusion, multiple landmarks, objects in variable scales, etc.
We show that DELF outperforms the state-of-the-art global and local descriptors in the large-scale setting by significant margins.
Code and dataset can be found at the project webpage: \url{https://github.com/tensorflow/models/tree/master/research/delf}.
\end{abstract}



\section{Introduction} \label{sec:introduction}
Large-scale image retrieval is a fundamental task in computer vision, since it is directly related to various practical applications, \eg, object detection, visual place recognition, and product recognition.
The last decades have witnessed tremendous advances in image retrieval systems---from hand-crafted features and indexing algorithms \cite{Lowe2004,Sivic2003,Philbin07,Jegou2008} to, more recently, methods based on convolutional neural networks (CNNs) for global descriptor learning \cite{arandjelovic2015netvlad,radenovic2016cnn,gordo2016deep}.

\begin{figure}[t]
\begin{center}
   \includegraphics[width=1.02\linewidth]{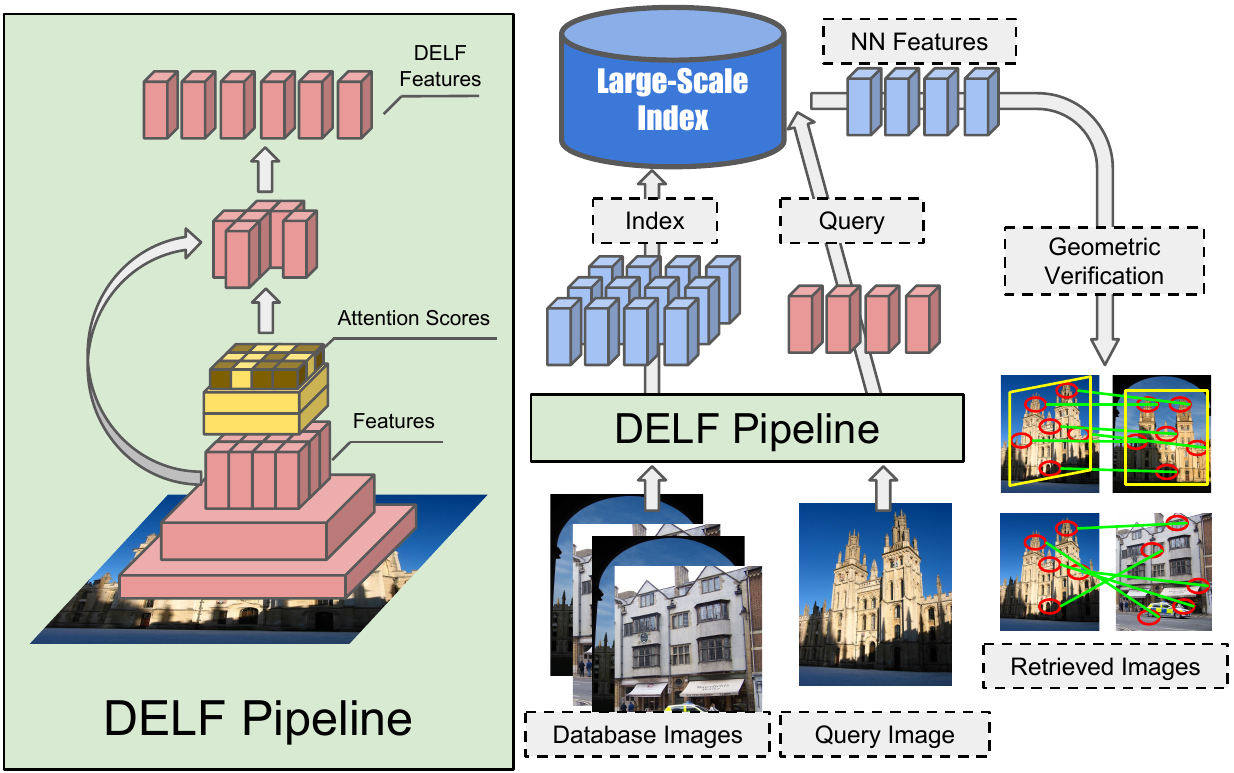}
\end{center}
\vspace{-12pt}                                                                                                                                                                                                           
   \caption{Overall architecture of our image retrieval system, using DEep Local Features (DELF) and attention-based keypoint selection.  On the left, we illustrate the pipeline for extraction and selection of DELF.  The portion highlighted in yellow represents an attention mechanism that is trained to assign high scores to relevant features and select the features with the highest scores.  Feature extraction and selection can be performed with a single forward pass using our model.  On the right, we illustrate our large-scale feature-based retrieval pipeline.  DELF for database images are indexed offline. The index supports querying by retrieving nearest neighbor (NN) features, which can be used to rank database images based on geometrically verified matches.}
\label{fig:key_fig}
\vspace{-10pt}                                                                                                                                                                                                           
\end{figure}

Despite the recent advances in CNN-based global descriptors for image retrieval in small or medium-size datasets~\cite{Philbin07,Philbin2008}, their performance may be hindered by a wide variety of challenging conditions observed in large-scale datasets, such as clutter, occlusion, and variations in viewpoint and illumination.
Global descriptors lack the ability to find patch-level matches between images.
As a result, it is difficult to retrieve images based on partial matching in the presence of occlusion and background clutter.
In a recent trend, CNN-based local features are proposed for patch-level matching \cite{han2015matchnet,zagoruyko2015learning,yi2016lift}.
However, these techniques are not optimized specifically for image retrieval since they lack the ability to detect semantically meaningful features, and show limited accuracy in practice.

Most existing image retrieval algorithms have been evaluated in small to medium-size datasets with few query images, \ie, only 55 in \cite{Philbin07,Philbin2008} and 500 in \cite{Jegou2008}, and the images in the datasets have limited diversity in terms of landmark locations and types.
Therefore, we believe that the image retrieval community can benefit from a large-scale dataset, comprising more comprehensive and challenging examples, to improve algorithm performance and  evaluation methodology by deriving more statistically meaningful results.
The main goal of this work is to develop a large-scale image retrieval system based on a novel CNN-based feature descriptor.
To this end, we first introduce a new large-scale dataset, Google-Landmarks, which contains more than 1M landmark images from almost 13K unique landmarks.
This dataset covers a wide area in the world, and is consequently more diverse and comprehensive than existing ones.
The query set is composed of an extra 100K images with diverse characteristics; in particular, we include images that have no match in the database, which makes our dataset more challenging.
This allows to assess the robustness of retrieval systems when queries do not necessarily depict landmarks.

We then propose a CNN-based local feature with attention, which is trained with weak supervision using image-level class labels only, without the need of object- and patch-level annotations.
This new feature descriptor is referred to as DELF (DEep Local Feature), and \figref{fig:key_fig} illustrates the overall procedure of feature extraction and image retrieval.
In our approach, the attention model is tightly coupled with the proposed descriptor; it reuses the same CNN architecture and generates feature scores using very little extra computation (in the spirit of recent advances in object detection \cite{ren2015faster}).
This enables the extraction of both local descriptors and keypoints via one forward pass over the network.
%
We show that our image retrieval system based on DELF achieves the state-of-the-art performance with significant margins compared to methods based on existing global and local descriptors.

\section{Related Work}

There are standard datasets commonly used for the evaluation of image retrieval techniques.
Oxford5k~\cite{Philbin07} has 5,062 building images captured in Oxford with 55 query images.
Paris6k~\cite{Philbin2008} is composed of 6,412 images of landmarks in Paris, and also has 55 query images.
These two datasets are often augmented with 100K distractor images from Flickr100k dataset~\cite{Philbin07}, which constructs Oxford105k and Paris106k datasets, respectively.
On the other hand, Holidays dataset~\cite{Jegou2008} provides 1,491 images including 500 query images, which are from personal holiday photos.
All these three datasets are fairly small, especially having a very small number of query images, which makes it difficult to generalize the performance tested in these datasets.
Although Pitts250k~\cite{torii13visual} is larger, it is specialized to visual places with repetitive patterns and may not be appropriate for the general image retrieval task.

Instance retrieval has been a popular research problem for more than a decade.
See \cite{zheng2016sift} for a recent survey.
Early systems rely on hand-crafted local features \cite{Lowe2004,bay2008speeded,buddemeier2012systems}, coupled with approximate nearest neighbor search methods using KD trees or vocabulary trees~\cite{beis1997shape,Nister}.
Still today, such feature-based techniques combined with geometric re-ranking provide strong performance when retrieval systems need to operate with high precision.

More recently, many works have focused on aggregation methods of local features, which include popular techniques such as VLAD \cite{Jegou2010} and Fisher Vector (FV) \cite{Jegou2012}.
The main advantage of such global descriptors is the ability to provide high-performance image retrieval with a compact index.

In the past few years, several global descriptors based on CNNs have been proposed to use pretrained~\cite{babenko2014neural,tolias2015particular} or learned networks~\cite{arandjelovic2015netvlad,radenovic2016cnn,gordo2016deep}.
These global descriptors are most commonly trained with a triplet loss, in order to preserve the ranking between relevant and irrelevant images.
Some retrieval algorithms using these CNN-based global descriptors make use of deep local features as a drop-in replacement for hand-crafted features in conventional aggregation techniques such as VLAD or FV \cite{yue2015exploiting,uricchio2015fisher}.
Other works have re-evaluated and proposed different feature aggregation methods using such deep local features \cite{babenko2015iccv,kalantidis2015cross}.

CNNs have also been used to detect, represent and compare local image features.
Verdie \etal \cite{verdie2015tilde} learned a regressor for repeatable keypoint detection.
Yi \etal \cite{moo2016learning} proposed a generic CNN-based technique to estimate the canonical orientation of a local feature and successfully deployed it to several different descriptors.
MatchNet~\cite{han2015matchnet} and DeepCompare~\cite{zagoruyko2015learning} have been proposed to jointly learn patch representations and associated metrics.
Recently, LIFT~\cite{yi2016lift} proposed an end-to-end framework to detect keypoints, estimate orientation, and compute descriptors.
Different from our work, these techniques are not designed for image retrieval applications since they do not learn to select semantically meaningful features.

Many visual recognition problems employ visual attention based on deep neural networks, which include object detection~\cite{zhou2016learning}, semantic segmentation~\cite{hong2015decoupled}, image captioning~\cite{xu2015show}, visual question answering~\cite{yang2016stacked}, etc.
However, visual attention has not been explored actively to learn visual features for image retrieval applications.

\section{Google-Landmarks Dataset}

\begin{figure}[t]
\vspace{3pt}
\begin{center}
\begin{subfigure}{1\linewidth}
   \includegraphics[width=1\linewidth]{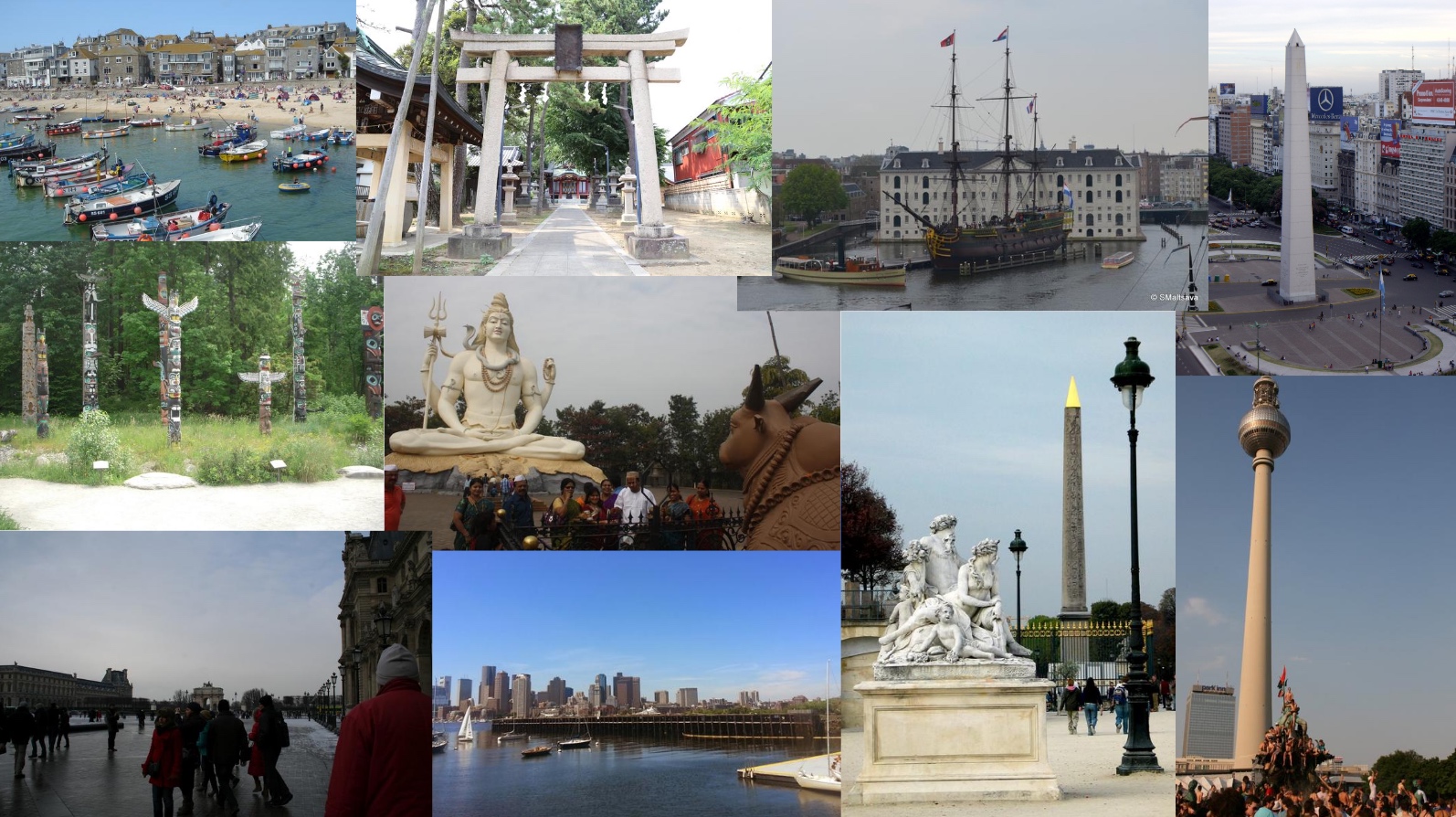} \vspace{-14pt}
   \caption{\footnotesize Sample database images} \vspace{1pt}
\end{subfigure}
\begin{subfigure}{1\linewidth}
   \includegraphics[width=1\linewidth]{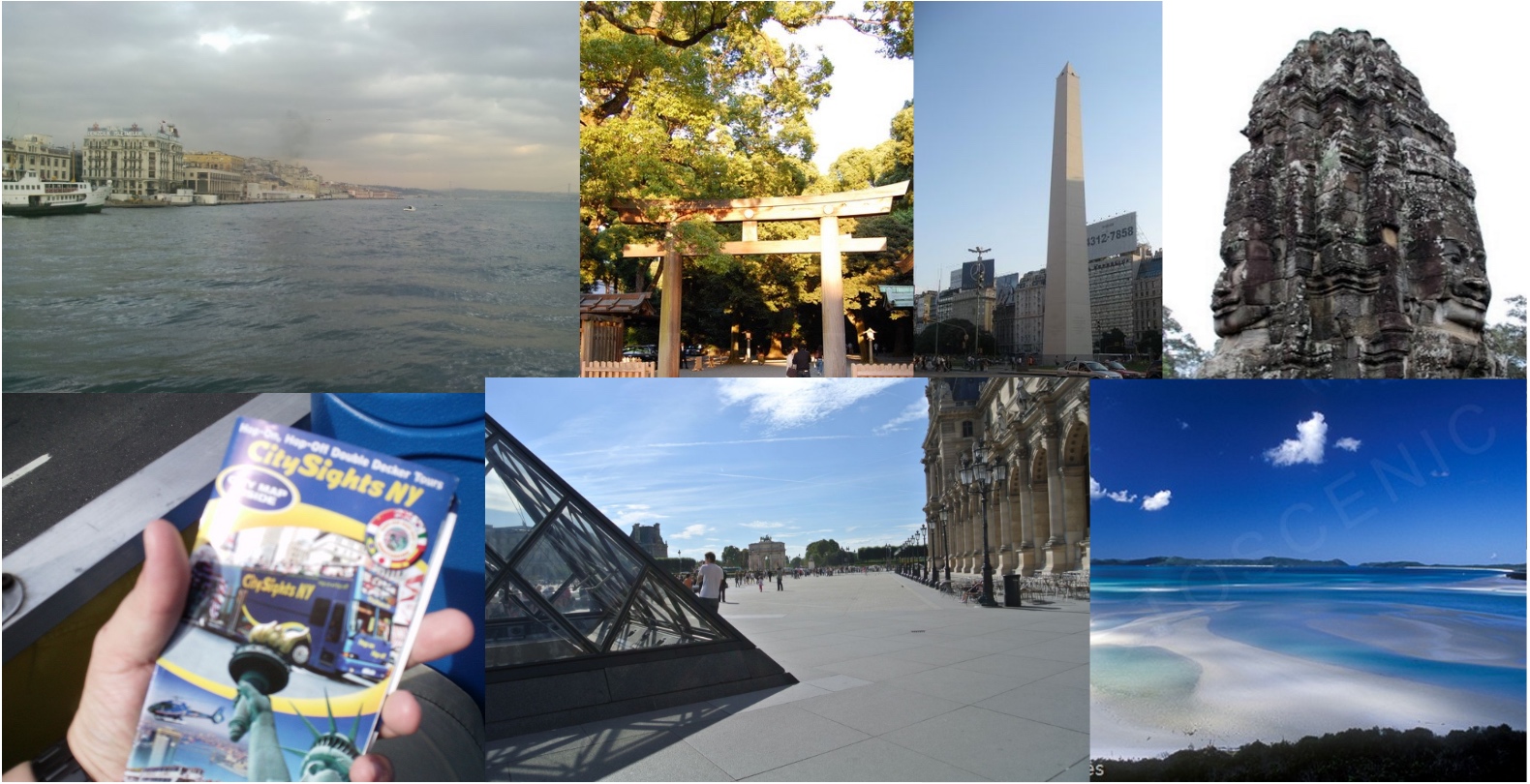} \vspace{-14pt}
   \caption{\footnotesize Sample query images} \vspace{1pt}
\end{subfigure}
\vspace{-8pt}          
\caption{Example database and query images from Google-Landmarks. They have a lot of variations and challenges including background clutter, small objects, and multiple landmarks.}
\vspace{-8pt}          
\label{fig:examples}                                                                                                                                                                                                 
\end{center}
\end{figure}

\begin{figure}[t]
\begin{center}
   \includegraphics[width=1\linewidth]{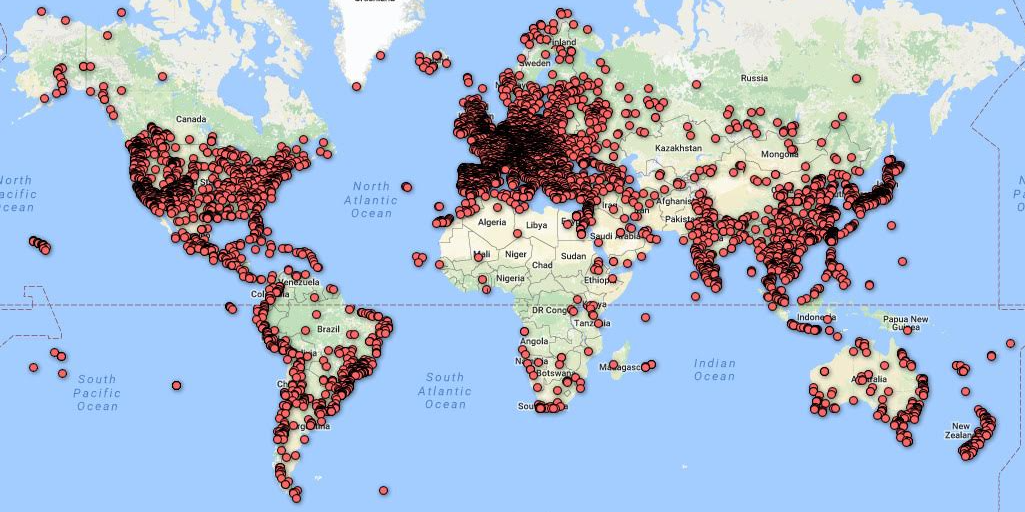}
\end{center}
	\vspace{-15pt}
	\caption{Image geolocation distribution of our Google-Landmarks dataset.  The landmarks are located in 4,872 cities in 187 countries.}
\label{fig:geolocation}
\vspace{-6pt}                                                                                                                                                                                                           
\end{figure}

Our dataset is constructed based on the algorithm described in \cite{zheng2009tour}.
Compared to the existing datasets for image retrieval~\cite{Philbin07,Philbin2008,Jegou2008}, the new dataset is much larger, contains diverse landmarks, and involves substantial challenges.
It contains $1,060,709$ images from $12,894$ landmarks, and $111,036$ additional query images.
The images in the dataset are captured at various locations in the world, and each image is associated with a GPS coordinate. 
Example images and their geographic distribution are presented in \figref{fig:examples} and \figref{fig:geolocation}, respectively.
While most images in the existing datasets are landmark-centric, which makes global feature descriptors work well, our dataset contains more realistic images with wild variations including foreground/background clutter, occlusion, partially out-of-view objects, etc.
In particular, since our query images are collected from personal photo repositories, some of them may not contain any landmarks and should not retrieve any image from the database.
We call these query images {\em distractors}, which play a critical role to evaluate robustness of algorithms to irrelevant and noisy queries.


We use visual features and GPS coordinates for ground-truth construction.
All images in the database are clustered using the two kinds of information, and we assign a landmark identifier to each cluster.
If physical distance between the location of a query image and the center of the cluster associated with the retrieved image is less than a threshold, we assume that the two images belong to the same landmark.
Note that ground-truth annotation is extremely challenging, especially considering the facts that it is hard to predefine what landmarks are, landmarks are not clearly noticeable sometimes, and there might be multiple instances in a single image.
%
Obviously, this approach for ground-truth construction is noisy due to GPS errors.
Also, photos can be captured from a large distance for some landmarks (\eg, Eiffel Tower, Golden Gate Bridge), and consequently the photo location might be relatively far from the actual landmark location.
However, we found very few incorrect annotations with the threshold of $25$km when checking a subset of data manually.
Even though there are few minor errors, it is not problematic, especially in relative evaluation, because algorithms are unlikely to be confused between landmarks anyway if their visual appearances are sufficiently discriminative.



\section{Image Retrieval with DELF} \label{sec:method}

Our large-scale retrieval system can be decomposed into four main blocks: (i) dense localized
feature extraction, (ii) keypoint selection, (iii) dimensionality reduction and 
(iv) indexing and retrieval.
This section describes DELF feature extraction and learning algorithm followed by our indexing and retrieval procedure in detail.

\subsection{Dense Localized Feature Extraction}
\label{sub:dense}
We extract dense features from an image by applying a fully convolutional network (FCN), which is constructed by using the feature extraction layers of a CNN trained with a classification loss.
We employ an FCN taken from the ResNet50~\cite{he2015deep} model, using the output of
the \textsf{conv4\_x} convolutional block.
To handle scale changes, we explicitly construct an image pyramid and apply the FCN for each level independently.
The obtained feature maps are regarded as a dense grid of local descriptors.
Features are localized based on their receptive fields, which can be computed by considering the configuration of convolutional and pooling layers of the FCN.
We use the pixel coordinates of the center of the receptive field as the feature location.
The receptive field size for the image at the original scale is $291\times 291$.
Using the image pyramid, we obtain features that describe image regions of different sizes.

\begin{figure}[t]
\begin{center}
   \includegraphics[width=0.9\linewidth]{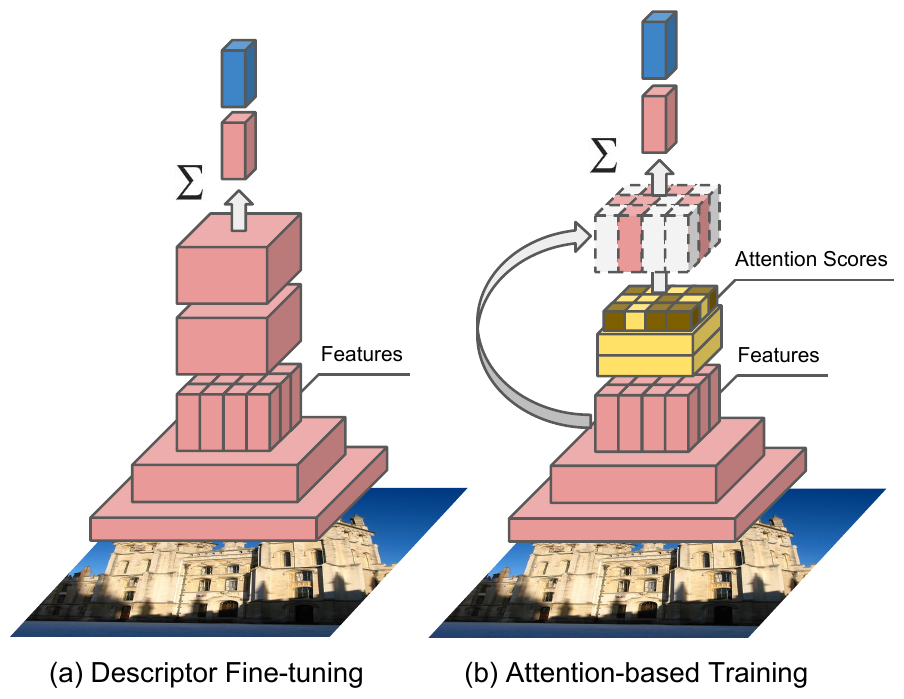}
\end{center}
	\vspace{-15pt}
	\caption{The network architectures used for training.}
\label{fig:train_model}
\vspace{-5pt}                                                                                                                                                                                                           
\end{figure}

We use the original ResNet50 model trained on ImageNet~\cite{russakovsky2015imagenet} as a baseline, and fine-tune for enhancing the discriminativeness of our local descriptors. 
Since we consider a landmark recognition application, we employ annotated datasets of landmark images~\cite{babenko2014neural} and train the network with a standard cross-entropy loss for image classification as illustrated in \figref{fig:train_model}(a).
The input images are initially center-cropped to produce square images and rescaled
to $250\times 250$.
Random $224\times 224$ crops are then used for training.
As a result of training, local descriptors implicitly learn representations that are more relevant for the landmark retrieval problem.
In this manner, neither object- nor patch-level labels are necessary to obtain improved local descriptors.

\subsection{Attention-based Keypoint Selection} \label{subsec:attention}

Instead of using densely extracted features directly for image retrieval, we design a technique to effectively select a subset of the features.
Since a substantial part of the densely extracted features are irrelevant to our recognition task and likely to add clutter, distracting the retrieval process, keypoint selection is important for both accuracy and computational efficiency of retrieval systems.


\subsubsection{Learning with Weak Supervision}

We propose to train a landmark classifier with attention to explicitly measure relevance scores for local feature descriptors.
To train the function, features are pooled by a weighted sum, where the weights are predicted by the attention network.
The training procedure is similar to the one described in \secref{sub:dense} including the loss function and datasets, and is illustrated in \figref{fig:train_model}(b), where the attention network is highlighted in yellow.
This generates an embedding for the whole input image, which is then used to train a softmax-based landmark classifier.

More precisely, we formulate the training as follows.
Denote by ${\mathbf{f}_n\in\mathrm{R}^d}, n=1,...,N$ the $d$-dimensional features to be learned jointly with the attention model.
Our goal is to learn a score function $\alpha(\mathbf{f}_n ; \theta)$ for each feature, where $\theta$ denotes the parameters of function $\alpha(\cdot)$.
The output logit $\mathbf{y}$ of the network is generated by a weighted sum of the feature vectors, which is given by
\begin{equation}
	\mathbf{y} = \mathbf{W} \left(
	\sum_{n}\alpha(\mathbf{f}_n ; \theta)\cdot\mathbf{f}_n
	\right),
\end{equation}
where $\mathbf{W}\in\mathrm{R}^{M \times d}$ represents the weights of the final fully-connected layer 
of the CNN trained to predict $M$ classes.

For training, we use cross entropy loss, which is given by
\begin{equation}
	\mathcal{L} =
		- \mathbf{y}^*
		\cdot \mathrm{log} \left(
		\frac{
			\mathrm{exp} \left(\mathbf{y}\right)
		}{
			\mathbf{1}^{\text T} \ \mathrm{exp}\left(\mathbf{y}\right)
		}
		\right),
\end{equation}
where $\mathbf{y}^*$ is ground-truth in one-hot representation and $\mathbf{1}$ is one vector.
The parameters in the score function $\alpha(\cdot)$ are trained by backpropagation, where the gradient is given by
\begin{equation}
\begin{split}
\frac
	{\partial \mathcal{L}}
	{\partial \theta}
	=
		\frac
			{\partial \mathcal{L}}
			{\partial \mathbf{y}}
		\sum_{n} 
			\frac
				{\partial \mathbf{y}}
				{\partial \alpha_n}
			\frac
				{\partial \alpha_n}
				{\partial \theta}
=
		\frac
			{\partial \mathcal{L}}
			{\partial \mathbf{y}}
		\sum_{n} 
			\mathbf{W}
			\hspace{0.03cm}
			\mathbf{f}_n
			\frac
				{\partial \alpha_n}
				{\partial \theta},
\end{split}
\end{equation}
where the backpropagation of the output score $\alpha_n \equiv \alpha(\mathbf{f}_n ; \theta)$ with respect to $\theta$ is same as the standard multi-layer perceptron.

We restrict $\alpha(\cdot)$ to be non-negative, to prevent it from learning negative weighting.
The score function is designed using a 2-layer CNN with a softplus~\cite{dugas2001incorporating} activation at the top.
For simplicity, we employ the convolutional filters of size $1\times 1$, which work well in practice.
Once the attention model is trained, it can be used to assess the relevance of features extracted by our model.
%

\subsubsection{Training Attention}

In the proposed framework, both the descriptors and the attention model are implicitly learned with image-level labels.
Unfortunately, this poses some challenges to the learning process.
While the feature representation and the score function can be trained jointly by
backpropagation, we found that this setup generates weak models in practice.
Therefore, we employ a two-step training strategy.
First, we learn descriptors with fine-tuning as described in \secref{sub:dense}, and then the score function is learned given the fixed descriptors.

Another improvement to our models is obtained by random image rescaling during attention training process.
This is intuitive, as the attention model should be able to generate effective scores for features at different scales.
In this case, the input images are initially center-cropped to produce square images, and rescaled
to $900\times 900$.
Random $720\times 720$ crops are then extracted and finally randomly scaled with a factor $\gamma \leq 1$.

\subsubsection{Characteristics}

One unconventional aspect of our system is that keypoint selection comes after descriptor extraction, which is different from the existing techniques (\eg, SIFT~\cite{Lowe2004} and LIFT~\cite{yi2016lift}), where keypoints are first detected and later described.
Traditional keypoint detectors focus on detecting keypoints repeatably under different imaging conditions, based only on their low-level characteristics.
However, for a high-level recognition task such as image retrieval, it is also critical to select keypoints that discriminate different object instances.
The proposed pipeline achieves both goals by training a model that encodes higher level semantics in the feature map, and learning to select discriminative features for the classification task. 
This is in contrast to recently proposed techniques for learning keypoint detectors, \ie, LIFT \cite{yi2016lift}, which collect training data based on SIFT matches.
Although our model is not constrained to learn invariances to pose and viewpoint, it implicitly learns to do so---similar to CNN-based image classification techniques.

\subsection{Dimensionality Reduction}

We reduce the dimensionality of selected features to obtain improved retrieval accuracy,
as common practice \cite{jegou2012negative}.
First, the selected features are $L_2$ normalized, and their dimensionality is reduced to 40 by PCA, which presents a good trade-off between compactness and discriminativeness.
Finally, the features once again undergo $L_2$ normalization.



\subsection{Image Retrieval System}

We extract feature descriptors from query and database images, where a predefined number of local features with the highest attention scores per image are selected.
Our image retrieval system is based on nearest neighbor search, which is implemented by a combination of KD-tree~\cite{bentley75multidimensional} and Product Quantization (PQ)~\cite{jegou2011product}.
We encode each descriptor to a 50-bit code using PQ, where each 40D feature descriptor is split into 10 subvectors with equal dimensions, and we identify $2^5$ centroids per subvector by $k$-means clustering to achieve 50-bit encoding. 
We perform asymmetric distance calculation, where the query descriptors are not encoded to improve the accuracy of nearest neighbor retrieval.
To speed up the nearest neighbor search, we construct an inverted index for descriptors, using a codebook of size 8K.
To reduce encoding errors, a KD-tree is used to partition each Voronoi cell, and a Locally Optimized Product Quantizer \cite{kalantidis2014locally} is employed for each subtree with fewer than 30K features.



When a query is given, we perform approximate nearest neighbor search for each local descriptor extracted from a query image. 
Then for the top $K$ nearest local descriptors retrieved from the index, we aggregate all the matches per database image. 
Finally, we perform geometric verification using RANSAC~\cite{Fischler1981} and employ the number of inliers as the score for retrieved images.
Many distractor queries are rejected by this geometric verification step because features from distractors may not be consistently matched with the ones from landmark images.

This pipeline requires less than 8GB memory to index 1 billion descriptors, which is sufficient to handle our large-scale landmark dataset.
The latency of the nearest neighbor search is less than 2 seconds using a single CPU under our experiment setup, where we soft-assign 5 centroids to each query and search up to 10K leaf nodes within each inverted index tree.


\section{Experiments} \label{sec:experiments}

This section mainly discusses the performance of DELF compared to existing global and local feature descriptors in our dataset.
In addition, we also show how DELF can be employed to achieve good accuracy in the existing datasets.

\subsection{Implementation Details}

\paragraph{Multi-scale descriptor extraction}
We construct image pyramids by using scales that are a $\sqrt{2}$ factor apart.
For the set of scales with range from $0.25$ to $2.0$, 7 different scales are used.
The size of receptive field is inversely proportional to the scale; for example, for the $2.0$ scale, the receptive field of the network covers $146 \times 146$ pixels.

\vspace{-10pt}
\paragraph{Training}
We employed landmarks dataset~\cite{babenko2014neural} for fine-tuning descriptors and training keypoint selection.
In the dataset, there are the ``full'' version, referred to as LF (after removal of overlapping classes with Oxf5k/Par6k, by \cite{gordo2016deep}), containing 140,372 images from 586 landmarks, and the ``clean'' version (LC) obtained by a SIFT-based matching procedure \cite{gordo2016deep}, with 35,382 images of 586 landmarks.
We use LF to train our attention model, and LC is employed to fine-tune the network for image retrieval.

\vspace{-10pt}
\paragraph{Parameters} 
We identify the top $K (= 60)$ nearest neighbors for each feature in a query and extract up to $1000$ local features from each image---each feature is $40$-dimensional.

\subsection{Compared Algorithms}
DELF is compared with several recent global and local descriptors.
Although there are various research outcomes related to image retrieval, we believe that the following methods are either relevant to our algorithm or most critical to evaluation due to their good performance.

\vspace{-10pt}
\paragraph{Deep Image Retrieval (DIR)~\cite{gordo2016deep}} 
This is a recent global descriptor that achieves the state-of-the-art performance in several existing datasets.  
DIR feature descriptors are $2,048$ dimensional and multi-resolution descriptors are used in all cases.  
We also evaluate with query expansion (QE), which typically improves accuracy in the standard datasets.
We use the released source code that implements the version with ResNet101~\cite{he2015deep}. 
For retrieval, a parallelized implementation of brute-force search is employed to avoid penalization by the error from approximate nearest neighbor search.

\vspace{-10pt}
\paragraph{siaMAC~\cite{radenovic2016cnn}} 
This is a recent global descriptor that obtains high performance in existing datasets.  
We use the released source code with parallelized implementation of brute-force search.
The CNN based on VGG16~\cite{Simonyan15} extracts $512$ dimensional global descriptor.  
We also experiment with query expansion (QE) as in DIR.

\vspace{-10pt}
\paragraph{CONGAS~\cite{buddemeier2012systems,neven2008image}} CONGAS is a $40$D hand-engineered local feature, which has been widely used for instance-level image matching and retrieval \cite{aradhye2009video2text,zheng2009tour}.  
This feature descriptor is extracted by collecting Gabor wavelet responses at the detected keypoint scale and orientation, and is known to have very similar performance and characteristic to other gradient-based local descriptors like SIFT. A Laplacian-of-Gaussian keypoint detector is used.

\vspace{-10pt}
\paragraph{LIFT} 
LIFT~\cite{yi2016lift} is a recently proposed feature matching pipeline, where keypoint detection, orientation estimation and keypoint description are jointly learned.  
Features are $128$ dimensional. 
We use the source code publicly available.

\subsection{Evaluation} 
\label{sub:metrics}
Image retrieval systems have typically been evaluated based on mean average precision (mAP), which is computed by sorting images in descending order of relevance per query and averaging AP of individual queries.
However, for datasets with distractor queries, such evaluation method is not representative since it is important to determine whether each image is relevant to the query or not.
In our case, the absolute retrieval score is used to estimate the relevance of each image.
For performance evaluation, we employ a modified version of precision (\textsc{Pre}) and recall (\textsc{Rec}) by considering all query images at the same time, which are given by 
\begin{align}
\textsc{Pre} = \frac{\sum_{q} | \mathcal{R}^\text{TP}_q |}{\sum_{q} |\mathcal{R}_q |}~~\text{and}~~
\textsc{Rec} = \sum_{q} | \mathcal{R}^\text{TP}_q |,
\label{eq:pr_def}
\end{align}
where $\mathcal{R}_q$ denotes a set of retrieved images for query $q$ given a threshold, and $\mathcal{R}_q^\text{TP} (\subseteq \mathcal{R}_q)$ is a set of true positives.
This is similar to the micro-AP metric introduced in \cite{perronnin2009}.
Note that in our case only the top-scoring image per landmark is considered in the final scoring.
We prefer unnormalized recall values, which present the number of retrieved true positives.
Instead of summarizing our result in a single number, we present a full precision-recall curve to inspect operating points with different retrieval thresholds.

\subsection{Quantitative Results}
\begin{figure}[t]
\begin{center}
\includegraphics[width=1\linewidth]{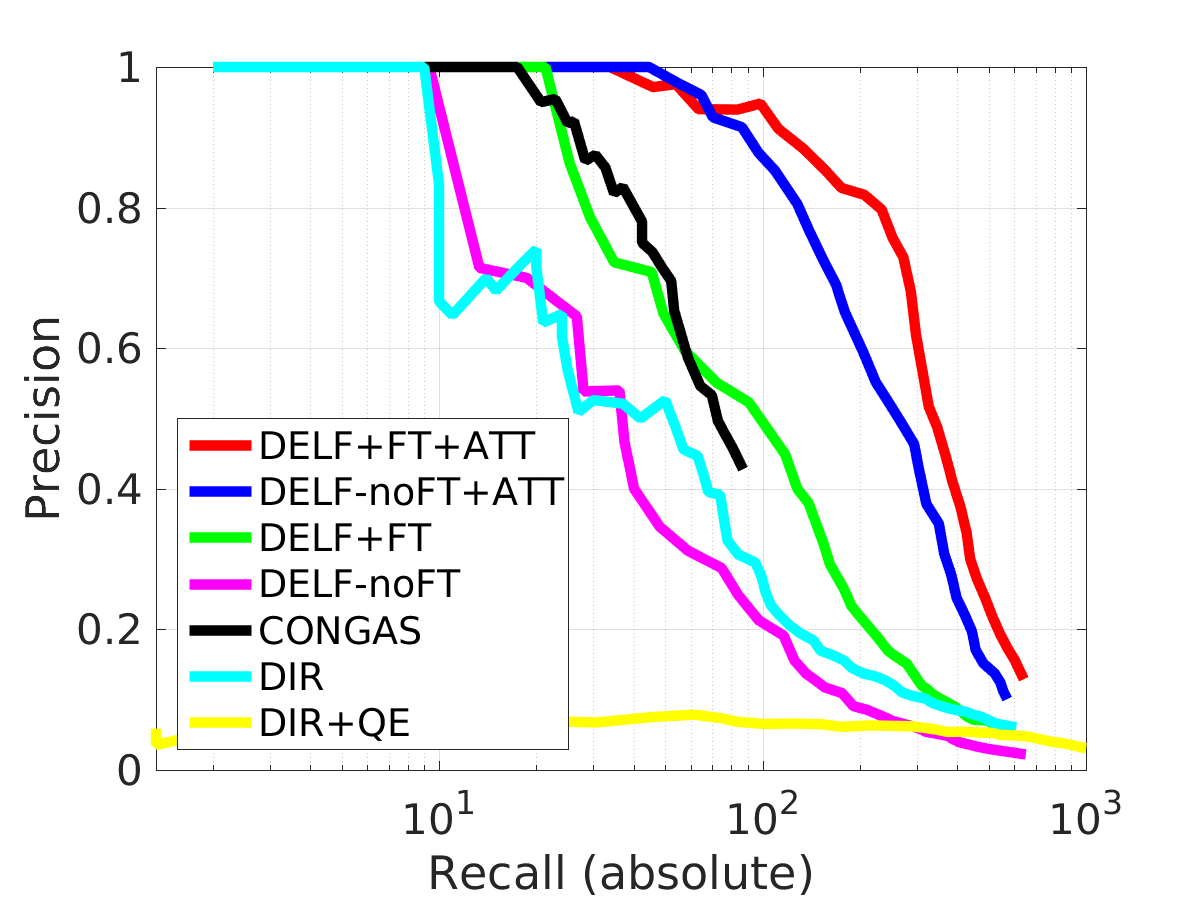}
\end{center}
\vspace{-15pt}                                                                                                                                                                                                           
   \caption{
Precision-recall curve for the large-scale retrieval experiment on the Google-Landmarks dataset, where recall is presented in absolute terms, as in Eq.~\eqref{eq:pr_def}.  DELF shows outstanding performance compared with existing global and local features.  Fine-tuning and attention model in DELF are critical to performance improvement.  The accuracy of DIR drops significantly with query expansion, due to many distractor queries in our dataset.}
\label{fig:large_scale_results}
\vspace{-5pt}
\end{figure}

\figref{fig:large_scale_results} presents the precision-recall curve of DELF (denoted by DELF+FT+ATT), compared to other methods.
The results of LIFT could not be shown because feature extraction is extremely slow and large-scale experiment is infeasible\footnote{LIFT feature extraction approximately takes 2 min/image using a GPU.}. 
%
DELF clearly outperforms all other techniques significantly.
Global feature descriptors, such as DIR, suffer in our challenging dataset.
In particular, due to a large number of distractors in the query set, DIR with QE degrades accuracy significantly.
CONGAS does a reasonably good job, but is still worse than DELF with substantial margin.

To analyze the benefit of fine-tuning and attention for image retrieval, we compare our full model (DELF+FT+ATT) with its variations: DELF-noFT, DELF+FT and DELF-noFT+ATT.
DELF-noFT means that extracted features are based on the pretrained CNN on ImageNet without fine-tuning and attention learning. 
DELF+FT denotes the model with fine-tuning but without attention modeling while DELF-noFT+ATT corresponds to the model without fine-tuning but using attention.
As illustrated in \figref{fig:large_scale_results}, both fine-tuning and attention modeling make substantial contributions to performance improvement.
In particular, note that the use of attention is more important than fine-tuning.
This demonstrates that the proposed attention layers effectively learn to select the most discriminative features for the retrieval task, even if the features are simply pretrained on ImageNet.

In terms of memory requirement, DELF, CONGAS and DIR are almost equally complex.
DELF and CONGAS adopt the same feature dimensionality and maximum number of features per image; they require approximately 8GB of memory.
DIR descriptors need 8KB per image, summing up to approximately 8GB to index the entire dataset.

\subsection{Qualitative Results} \label{sub:qualitative}

We present qualitative results to illustrate performance of DELF compared to two competitive algorithms based on global and local features---DIR and CONGAS, respectively.
Also, we analyze our attention-based keypoint detection algorithm by visualization.

\begin{figure}[t]
\begin{center}
   \makebox[0.32\linewidth]{ \includegraphics[height=2.0cm, width=0.32\linewidth]{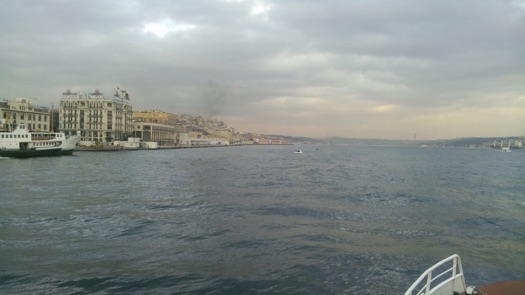}}
   \makebox[0.32\linewidth]{ \includegraphics[height=2.0cm, width=0.32\linewidth]{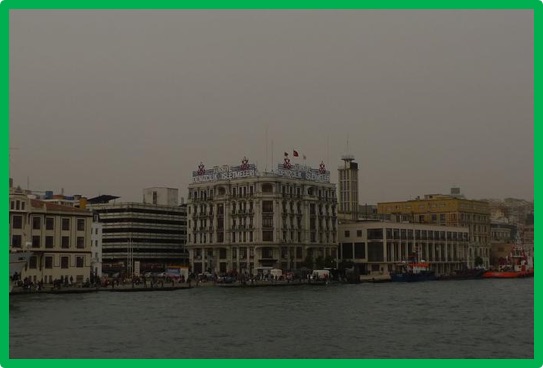}}
   \makebox[0.32\linewidth]{ \includegraphics[height=2.0cm, width=0.32\linewidth]{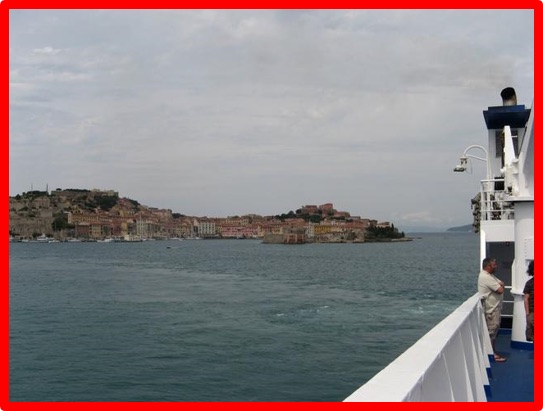}}\\   
   \vspace{0.05cm}   
   \makebox[0.32\linewidth]{ \includegraphics[height=2.0cm, width=0.32\linewidth]{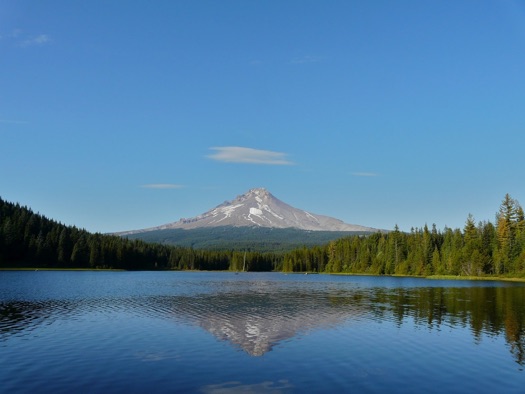}}
   \makebox[0.32\linewidth]{ \includegraphics[height=2.0cm, width=0.32\linewidth]{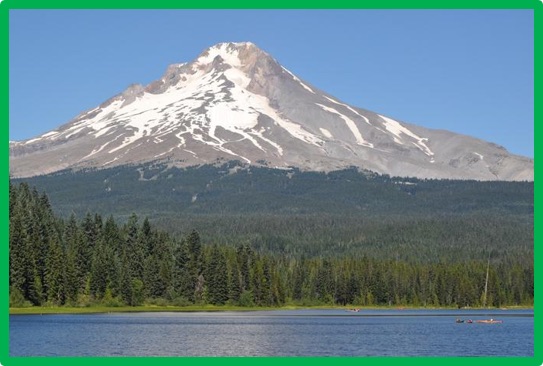}}
   \makebox[0.32\linewidth]{ \includegraphics[height=2.0cm, width=0.32\linewidth]{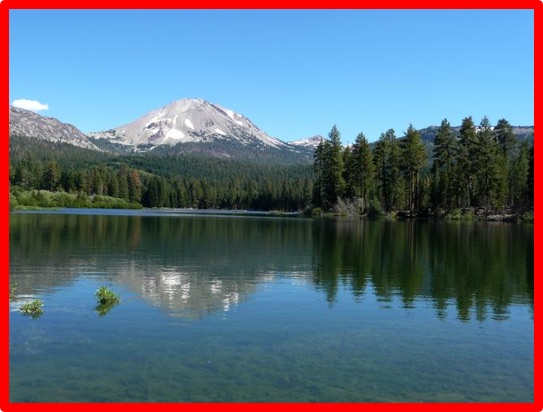}} \\
   \vspace{0.05cm}   
   \makebox[0.32\linewidth]{ \includegraphics[height=2.0cm, width=0.32\linewidth]{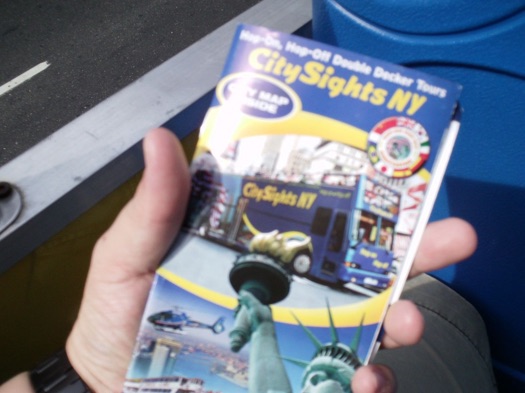}}
   \makebox[0.32\linewidth]{ \includegraphics[height=2.0cm, width=0.32\linewidth]{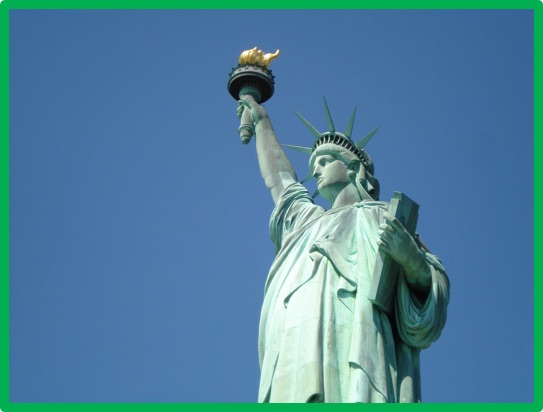}}
   \makebox[0.32\linewidth]{ \includegraphics[height=2.0cm, width=0.32\linewidth]{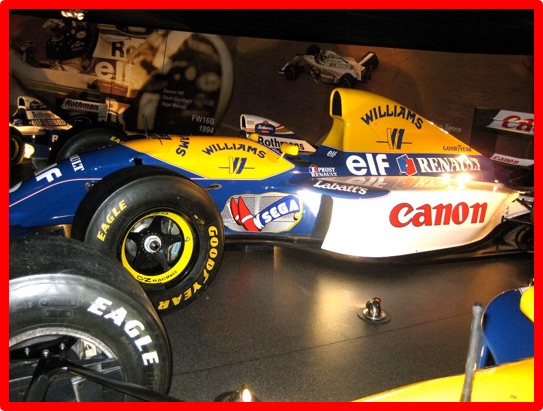}} \\
   \vspace{0.05cm}
   \makebox[0.32\linewidth]{ \includegraphics[height=2.0cm, width=0.32\linewidth]{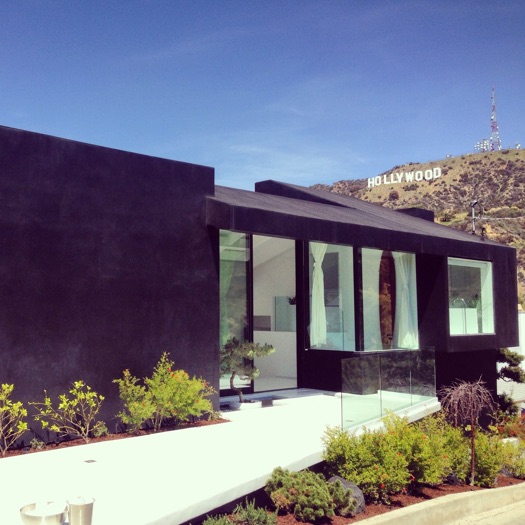}}
   \makebox[0.32\linewidth]{ \includegraphics[height=2.0cm, width=0.32\linewidth]{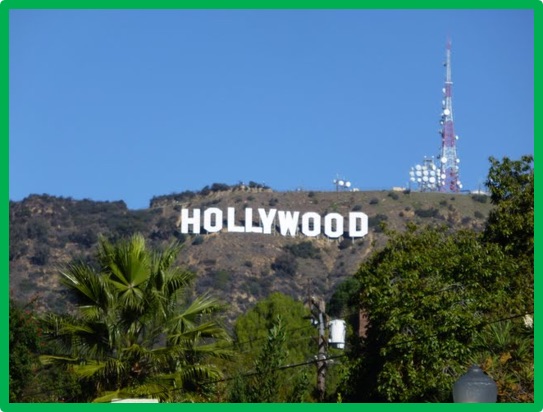}}
   \makebox[0.32\linewidth]{ \includegraphics[height=2.2cm]{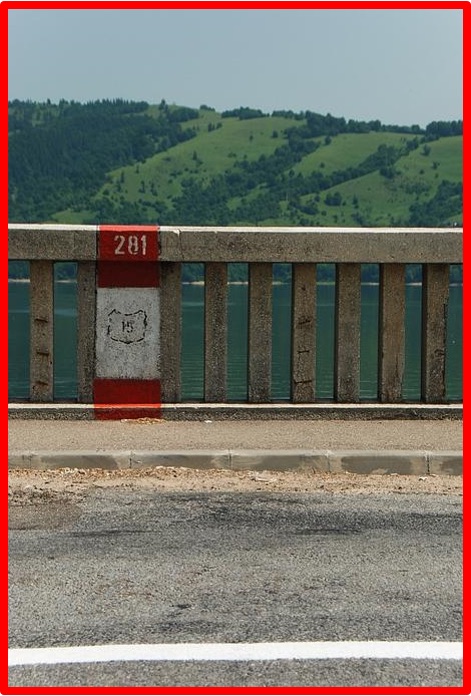}}\\     
   \vspace{0.05cm}
   \makebox[0.32\linewidth]{ \includegraphics[height=2.2cm]{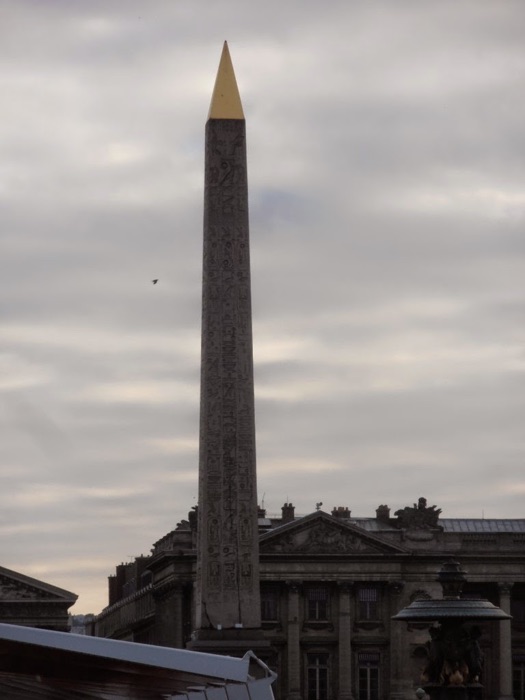}}
   \makebox[0.32\linewidth]{ \includegraphics[height=2.2cm]{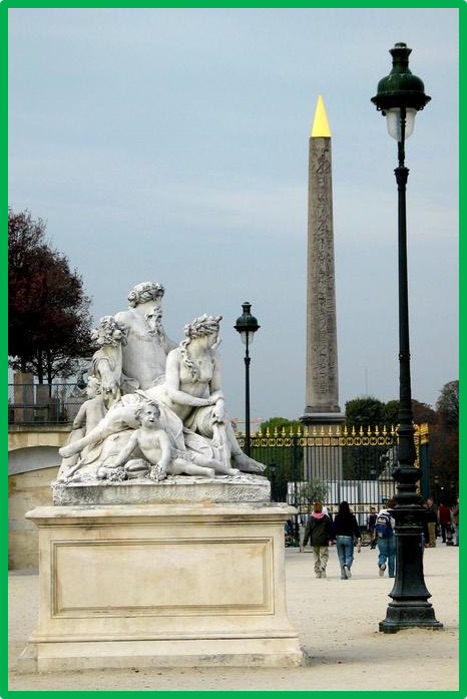}}
   \makebox[0.32\linewidth]{ \includegraphics[width=0.32\linewidth]{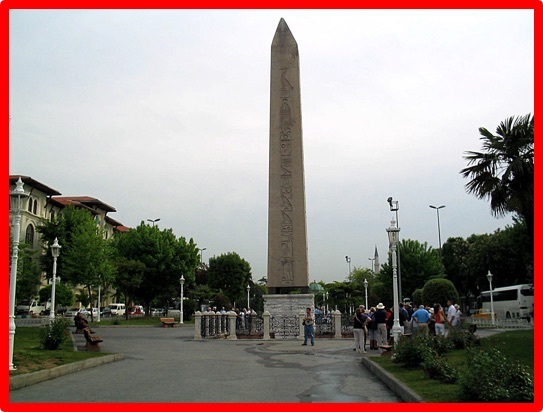}} \\
   \vspace{0.05cm}
   \makebox[0.32\linewidth]{ \includegraphics[height=2.0cm, width=0.32\linewidth]{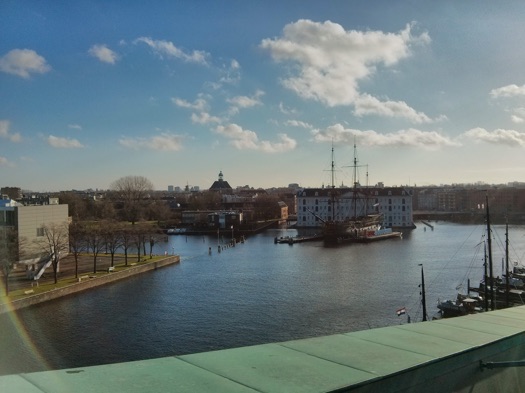}}
   \makebox[0.32\linewidth]{ \includegraphics[height=2.0cm, width=0.32\linewidth]{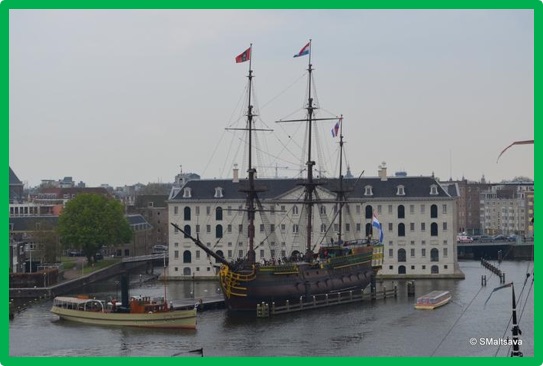}}
   \makebox[0.32\linewidth]{ \includegraphics[height=2.0cm, width=0.32\linewidth]{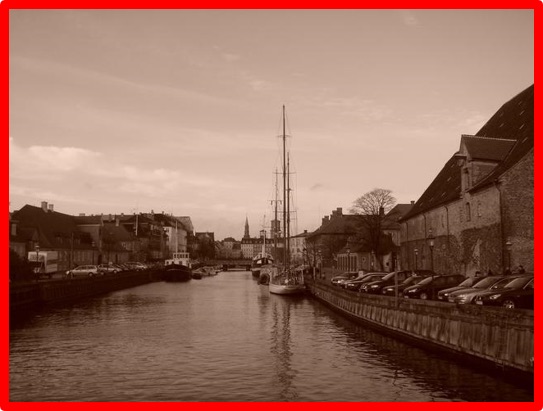}}\\
   \vspace{-0.1cm}
\centerline{\footnotesize (a) \hspace{2.3cm} (b) \hspace{2.3cm} (c)}
   \vspace{-0.5cm}   
\end{center}
	\caption{Examples where DELF+FT+ATT outperforms DIR: (a) query image, (b) top-1 image of DELF+FT+ATT, (c) top-1 image of DIR.  The green borders denote correct results while  the red ones mean incorrect retrievals.  Note that DELF deals with clutter in query and database images and small landmarks effectively.}
\label{fig:qualitative1}
\vspace{-5pt}
\end{figure}

\vspace{-5pt}
\paragraph{DELF~vs.~DIR}
\figref{fig:qualitative1} shows retrieval results, where DELF outperforms DIR.
DELF obtains matches between specific local regions in images, which helps significantly to find the same object in different imaging conditions.
Common failure cases of DIR happen when the database contains similar objects or scenes, \eg, obelisks, mountains, harbors, as illustrated in \figref{fig:qualitative1}.
In many cases, DIR cannot distinguish these specific objects or scenes; although it finds semantically similar images, they often do not correspond to the instance of interest.
Another weakness of DIR and other global descriptors is that they are not good at identifying small objects of interest.
\begin{figure}[t]
\begin{center}
   \makebox[0.32\linewidth]{ \includegraphics[height=2.0cm, width=0.32\linewidth]{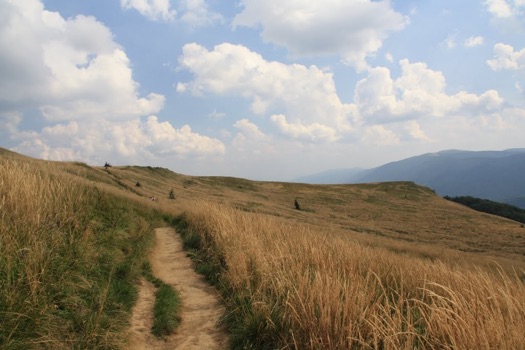}}
   \makebox[0.32\linewidth]{ \includegraphics[height=2.0cm, width=0.32\linewidth]{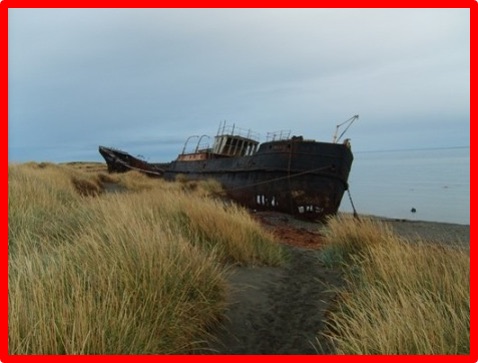}}
   \makebox[0.32\linewidth]{ \includegraphics[height=2.0cm, width=0.32\linewidth]{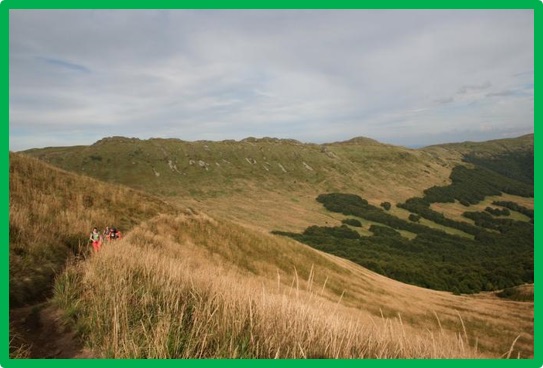}}\\
   \vspace{0.05cm}
   \makebox[0.32\linewidth]{ \includegraphics[height=2.2cm]{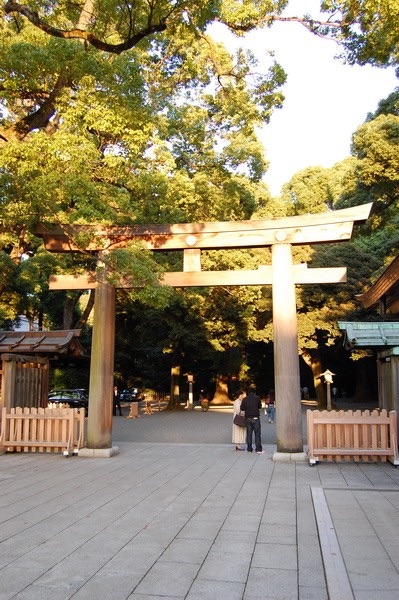}}
   \makebox[0.32\linewidth]{ \includegraphics[height=2.2cm]{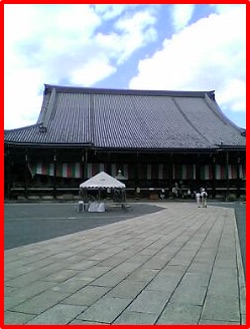}}
   \makebox[0.32\linewidth]{ \includegraphics[height=2.0cm, width=0.32\linewidth]{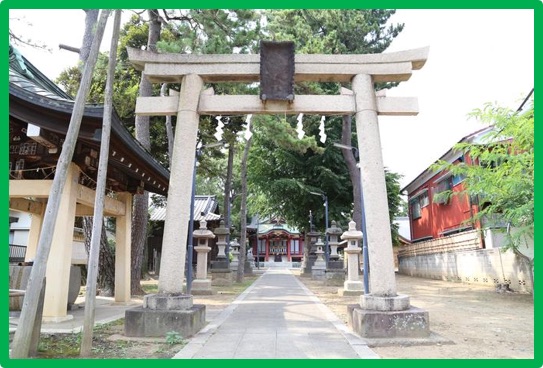}}\\   
   \vspace{0.05cm}   
   \makebox[0.32\linewidth]{ \includegraphics[height=2.0cm, width=0.32\linewidth]{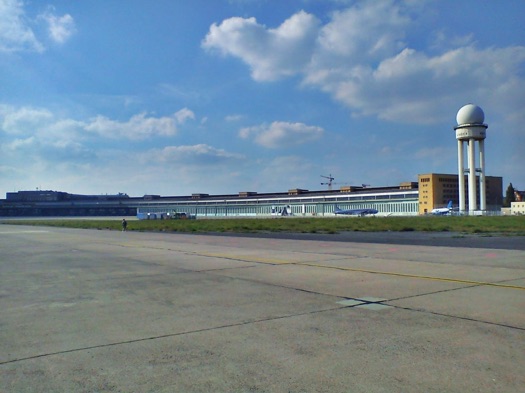}}
   \makebox[0.32\linewidth]{ \includegraphics[height=2.0cm, width=0.32\linewidth]{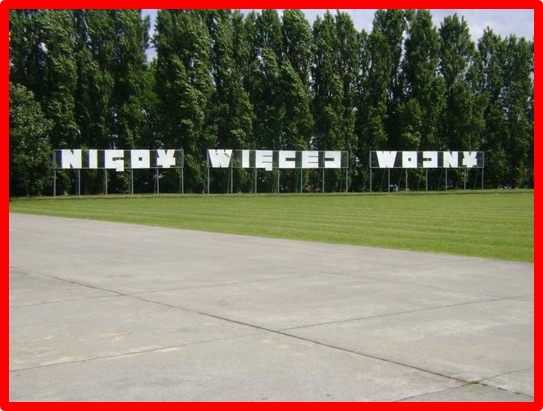}}
   \makebox[0.32\linewidth]{ \includegraphics[height=2.0cm, width=0.32\linewidth]{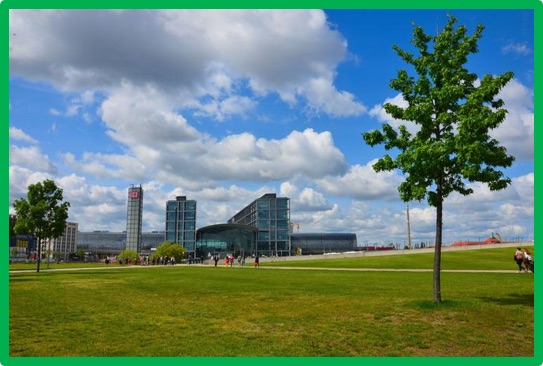}}\\
   \vspace{-0.1cm}   
\centerline{\footnotesize (a) \hspace{2.3cm} (b) \hspace{2.3cm} (c)}
   \vspace{-0.6cm}   
\end{center}
	\caption{Examples where DIR outperforms DELF+FT+ATT: (a) query image, (b) top-1 image of DELF+FT+ATT, (c) top-1 image of DIR.  The green and red borders denotes correct and incorrect results, respectively.}
	\vspace{-5pt}
\label{fig:qualitative2}
\end{figure}
\begin{figure*}[t]
\begin{center}
   \includegraphics[width=0.323\linewidth]{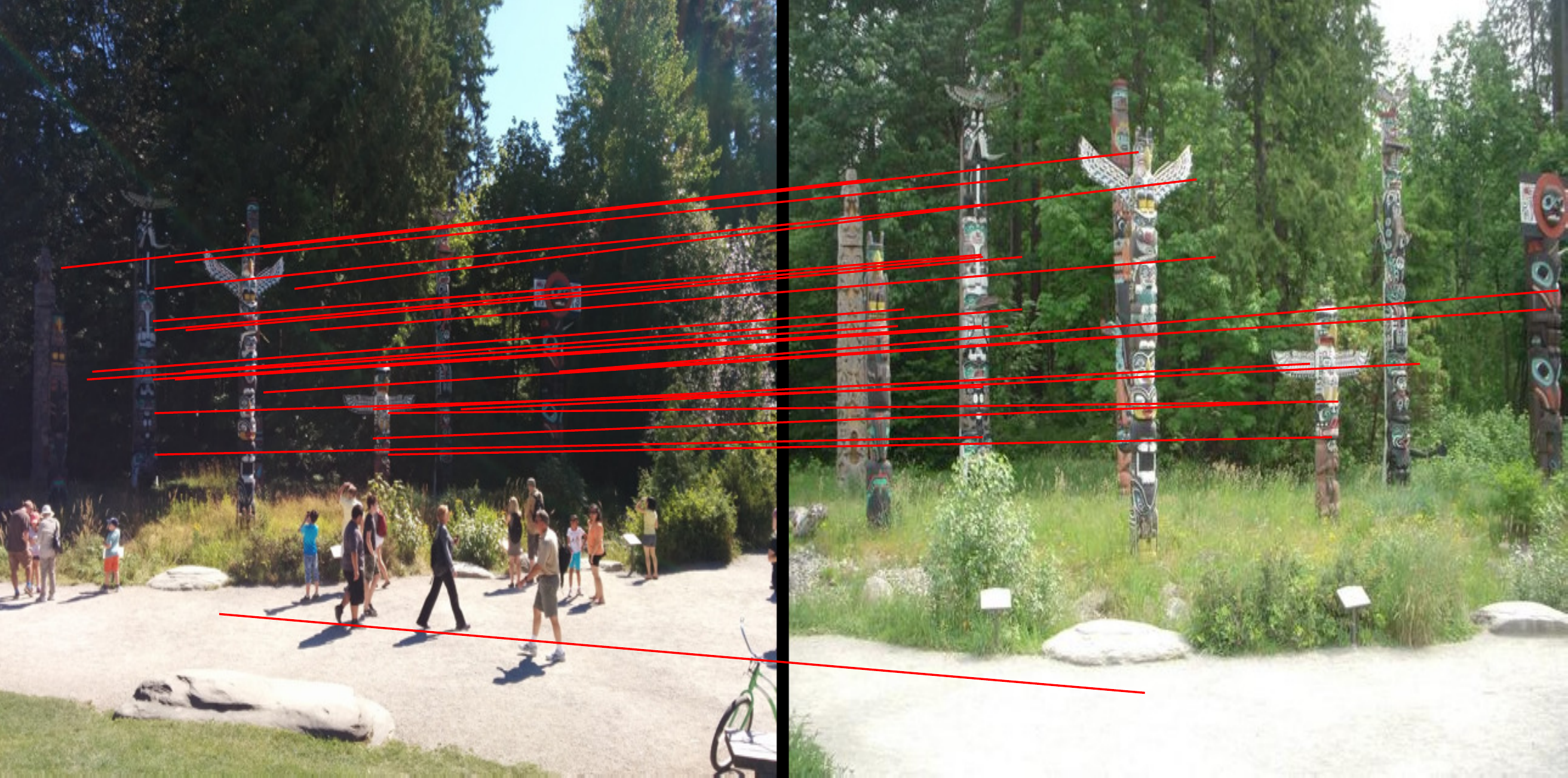}
   \includegraphics[width=0.323\linewidth]{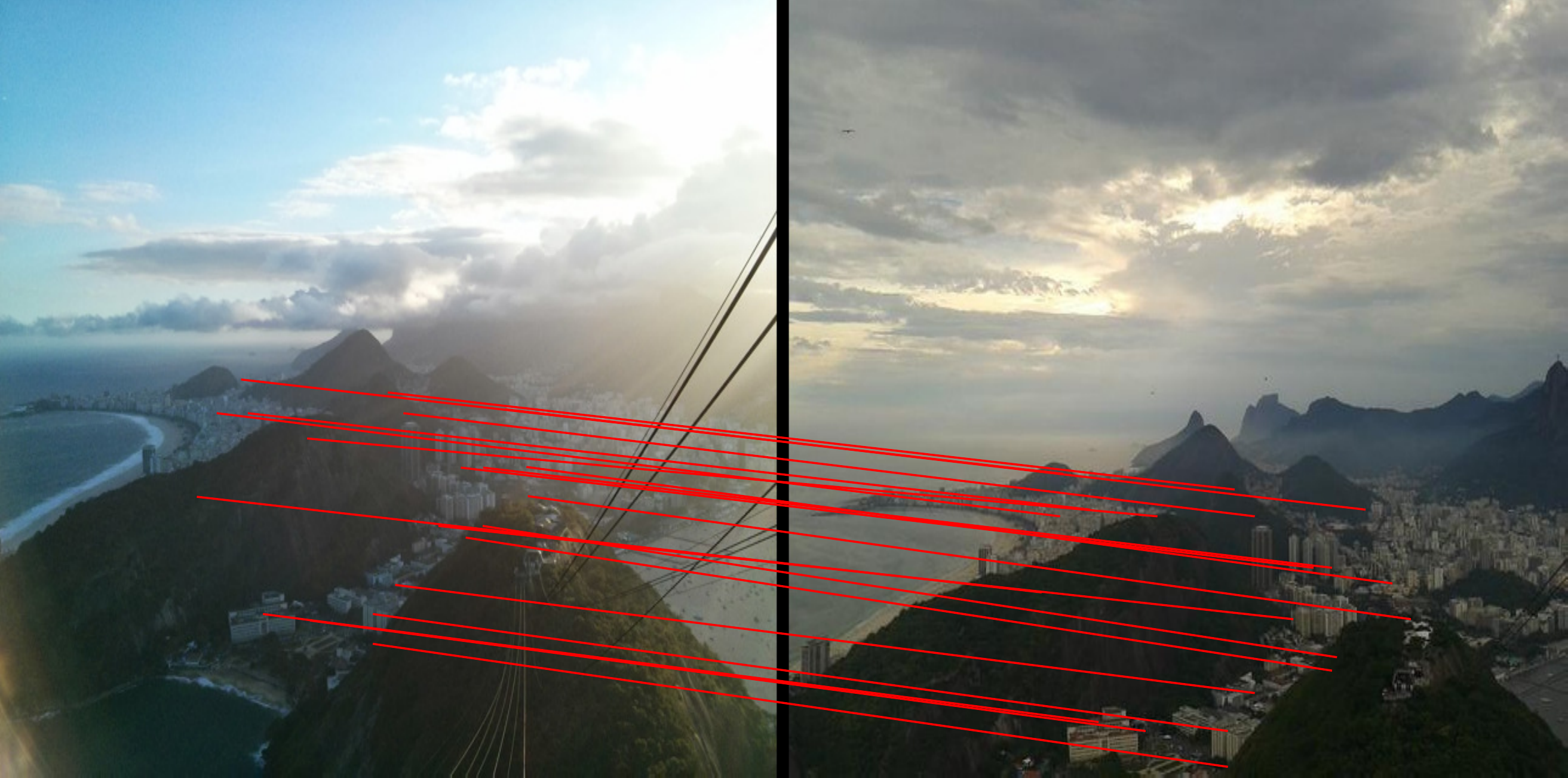}
   \includegraphics[width=0.323\linewidth]{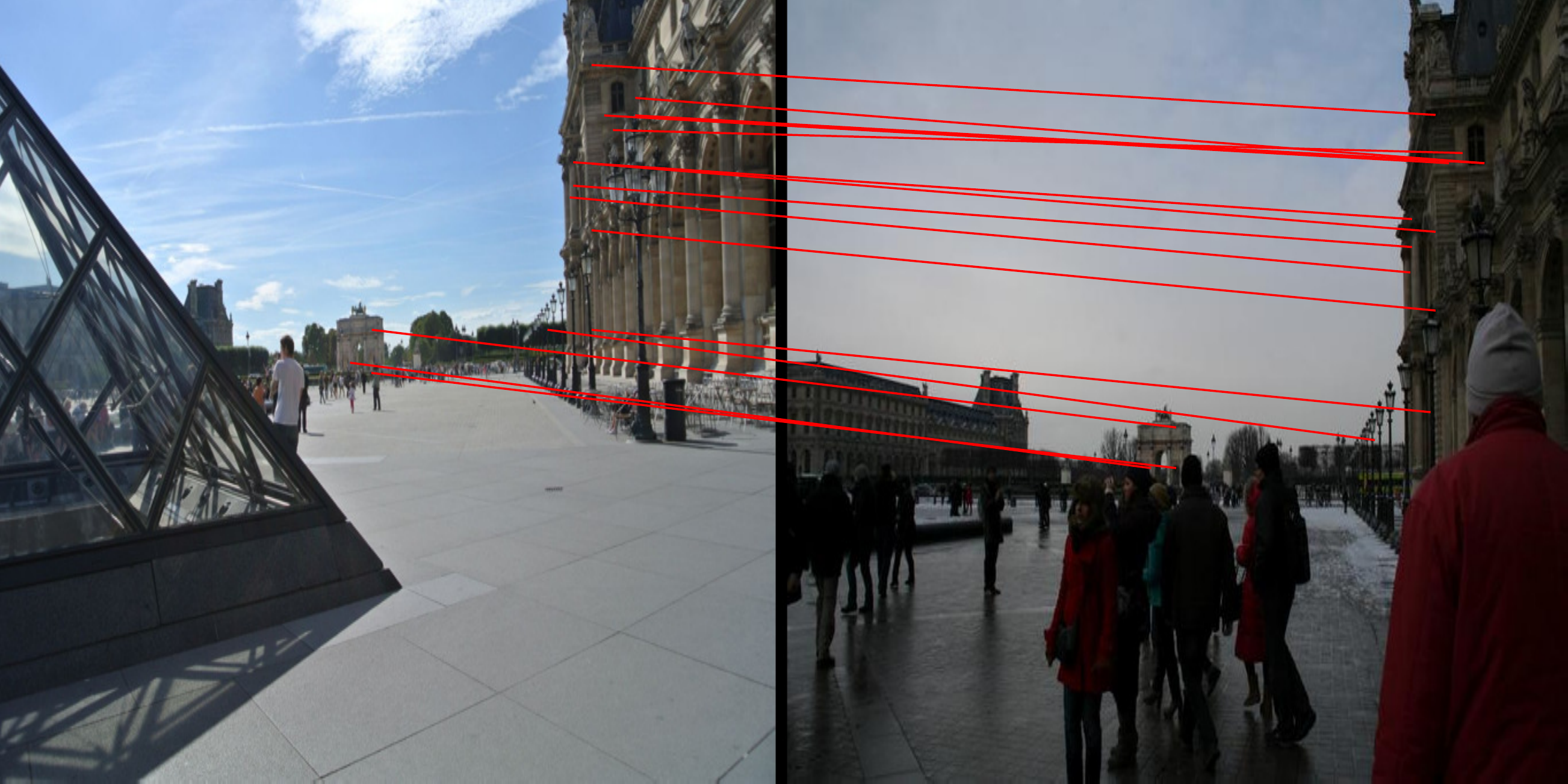} \\
   \vspace{0.03cm}      
   \includegraphics[width=0.323\linewidth]{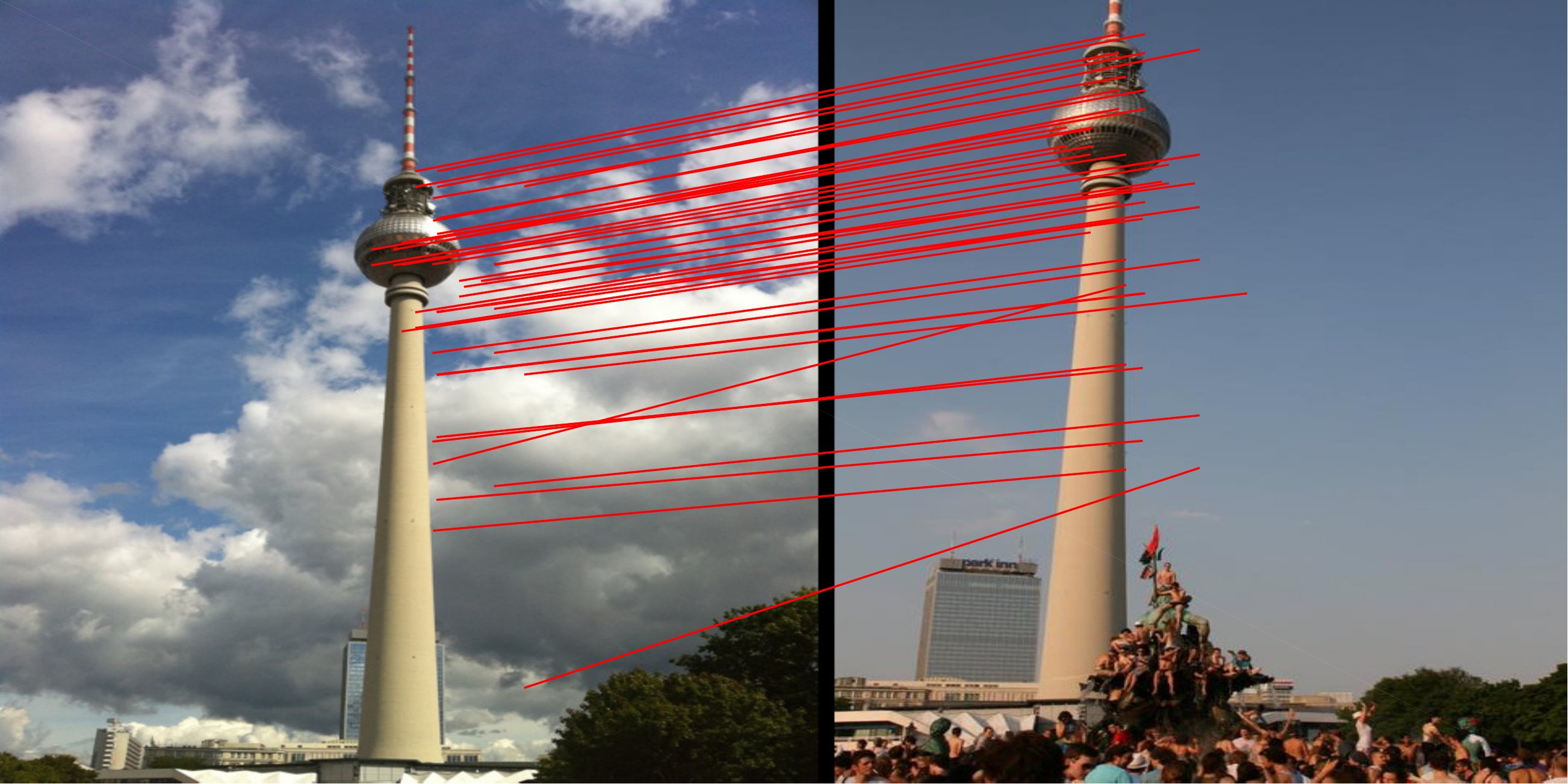} 
   \includegraphics[width=0.323\linewidth]{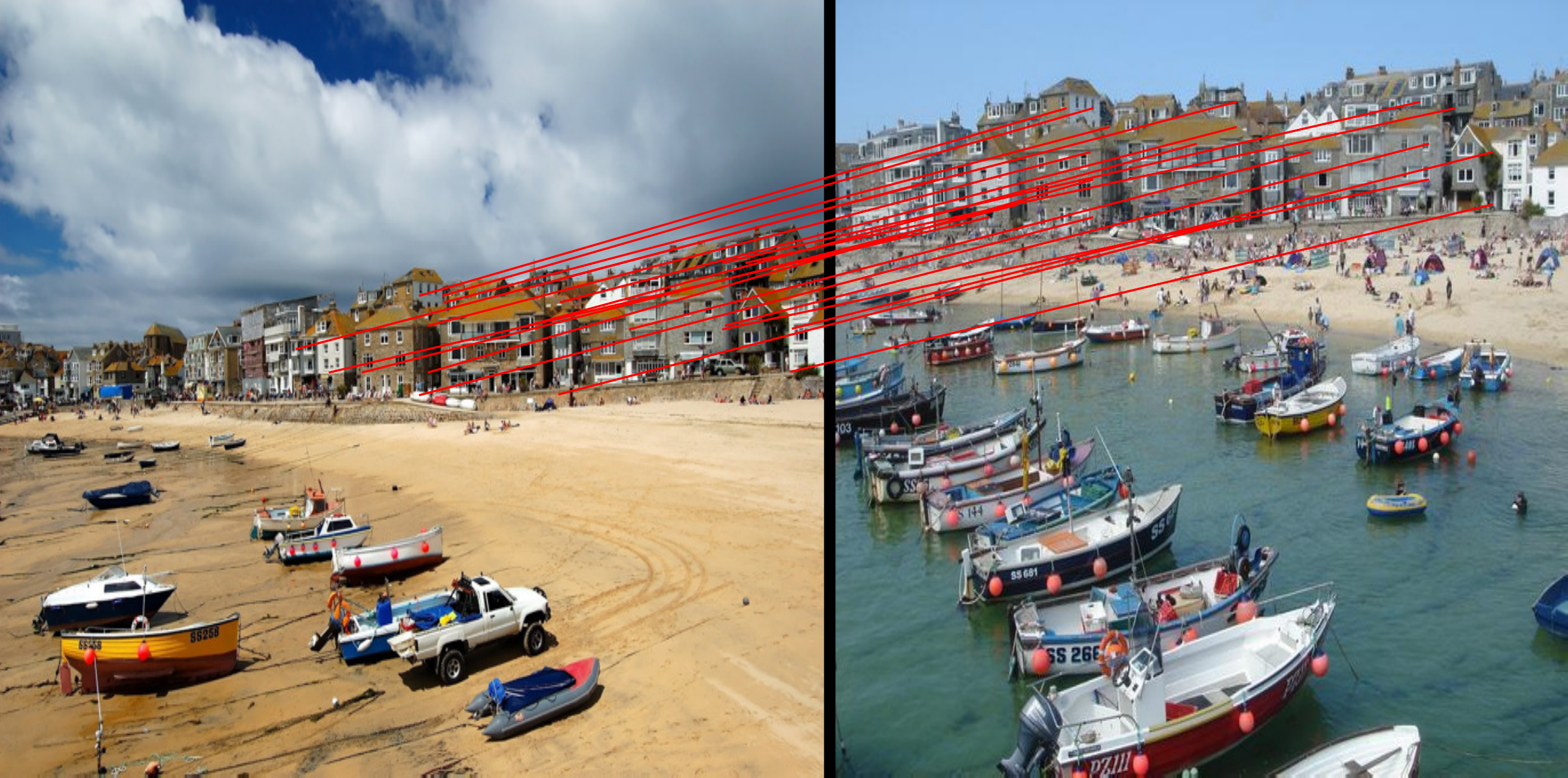}
   \includegraphics[width=0.323\linewidth]{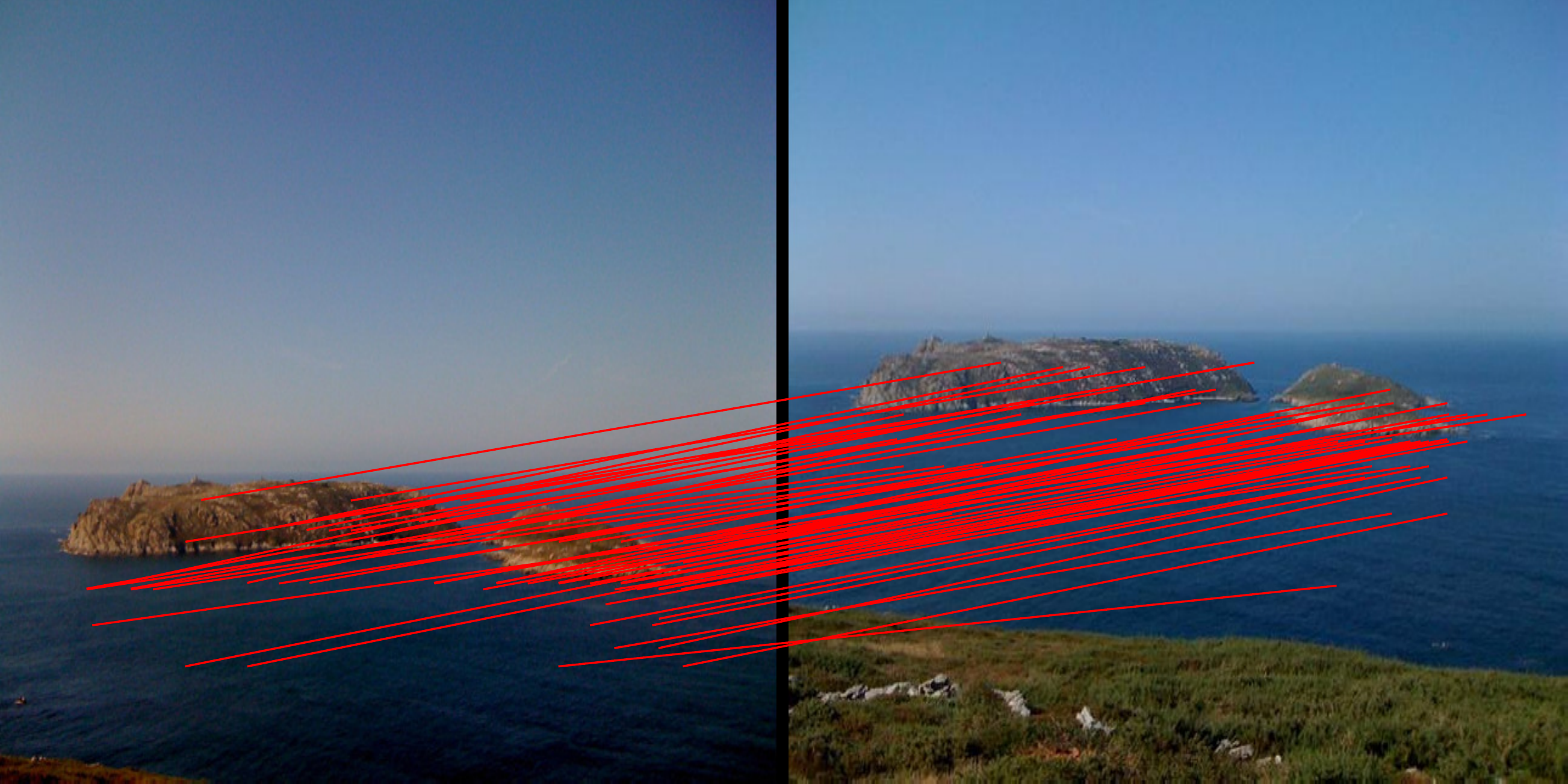} \\
   \vspace{-0.5cm}   
\end{center}
	\caption{Visualization of feature correspondences between images in query and database using DELF+FT+ATT.  For each pair, query and database images are presented side-by-side.  DELF successfully matches landmarks and objects in challenging environment including partial occlusion, distracting objects, and background clutter.  Both ends of the red lines denote the centers of matching features.  Since the receptive fields are fairly large, the centers may be located outside landmark object areas.  For the same queries, CONGAS fails to retrieve any image.}
	\vspace{-0.2cm}
\label{fig:feat_matches}
\end{figure*}
\figref{fig:qualitative2} shows the cases that DIR outperforms DELF.
While DELF is able to match localized patterns across different images, this leads to errors when the floor tiling or vegetation is similar across different landmarks.

\begin{figure}[!h]
\vspace{0.1cm}
\begin{center}
   \includegraphics[width=0.242\linewidth]{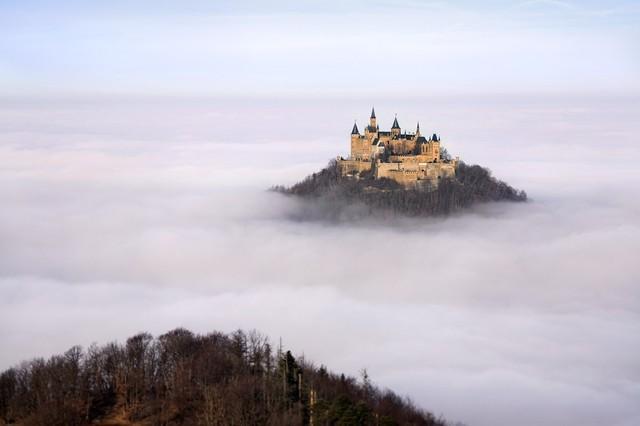}
   \includegraphics[width=0.242\linewidth]{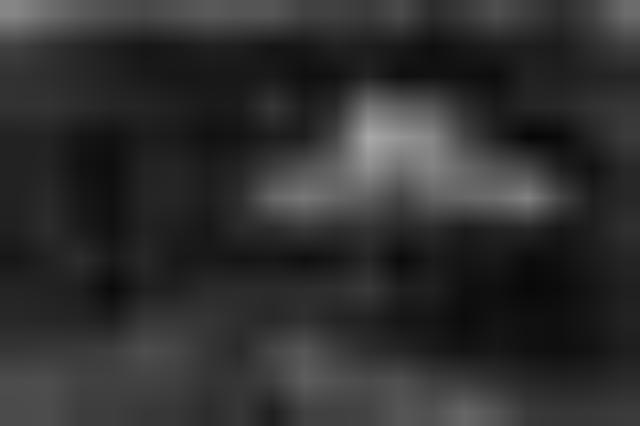}
   \includegraphics[width=0.242\linewidth]{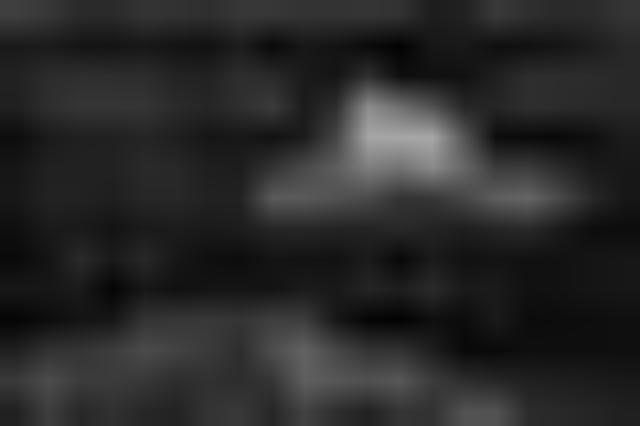}
   \includegraphics[width=0.242\linewidth]{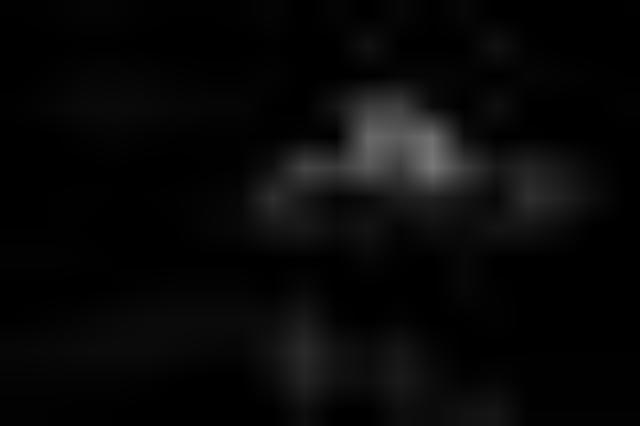}\\
	\vspace{0.05cm}
   \includegraphics[width=0.242\linewidth]{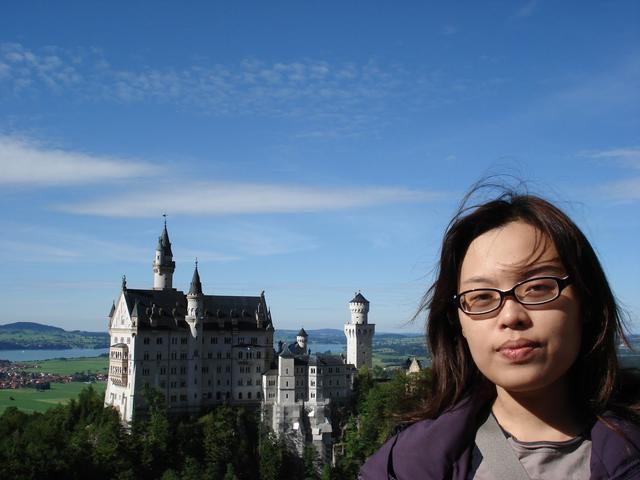}
   \includegraphics[width=0.242\linewidth]{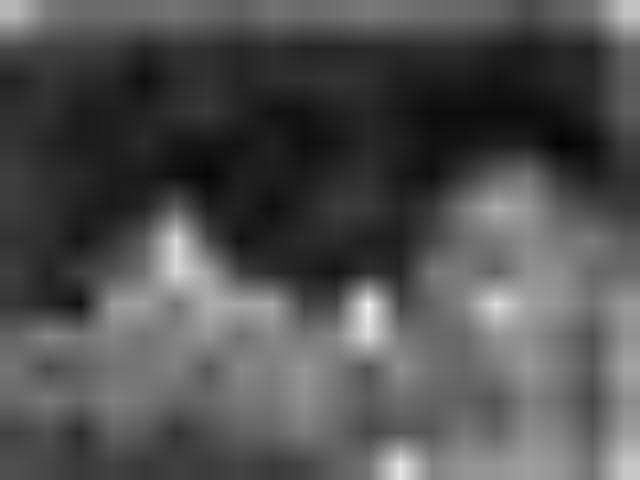}
   \includegraphics[width=0.242\linewidth]{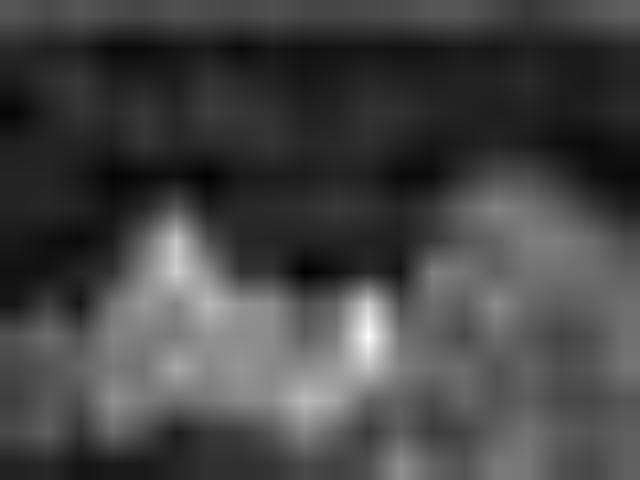}
   \includegraphics[width=0.242\linewidth]{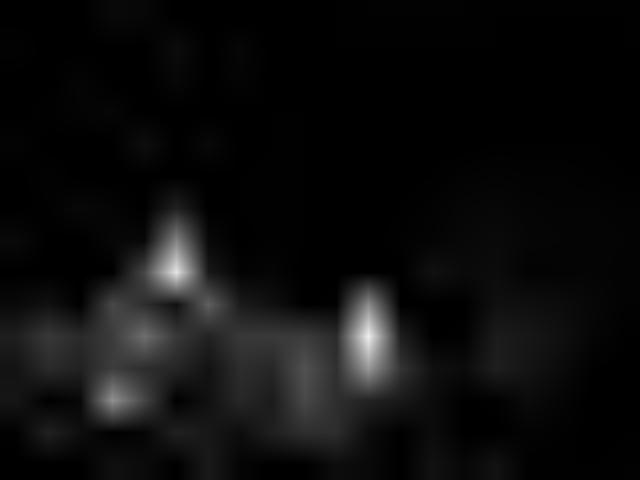}\\
	\vspace{0.05cm}
   \includegraphics[width=0.242\linewidth]{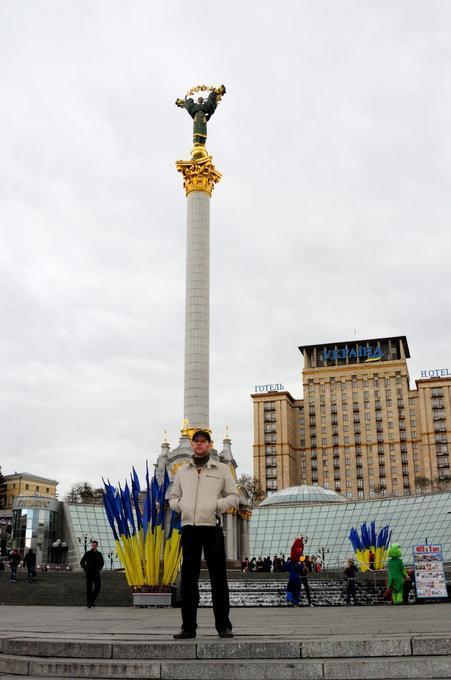}
   \includegraphics[width=0.242\linewidth]{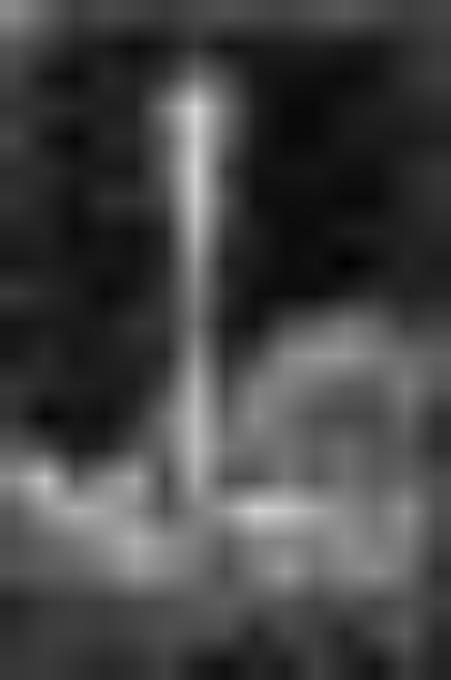}
   \includegraphics[width=0.242\linewidth]{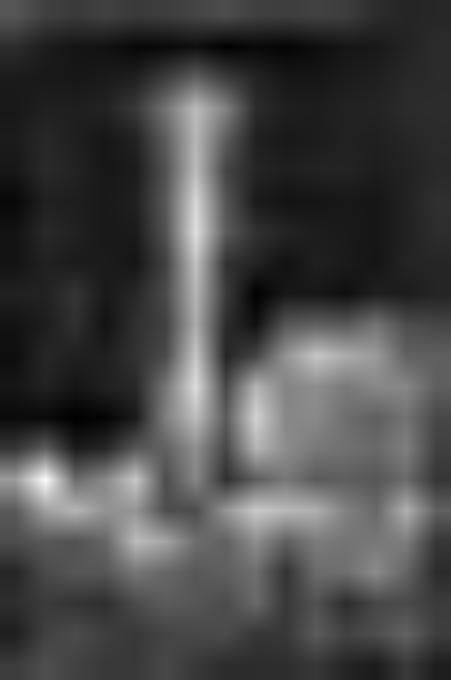}
   \includegraphics[width=0.242\linewidth]{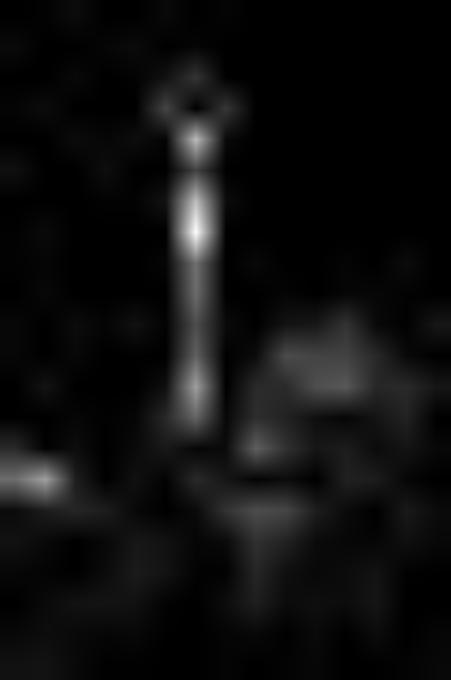}\\
	\vspace{0.05cm}
   \includegraphics[width=0.242\linewidth]{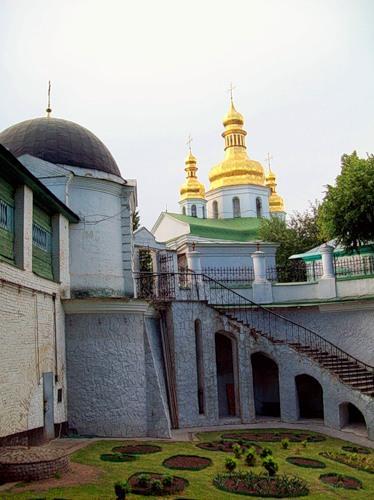}
   \includegraphics[width=0.242\linewidth]{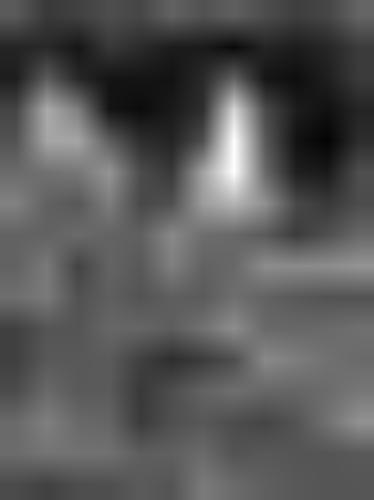}
   \includegraphics[width=0.242\linewidth]{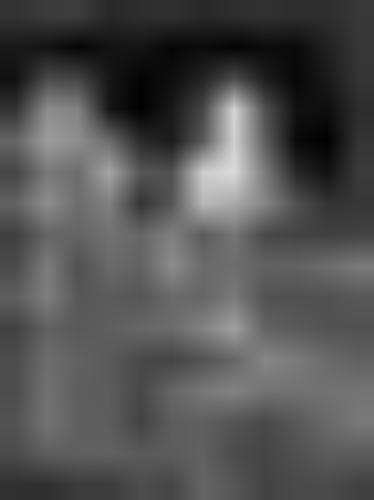}
   \includegraphics[width=0.242\linewidth]{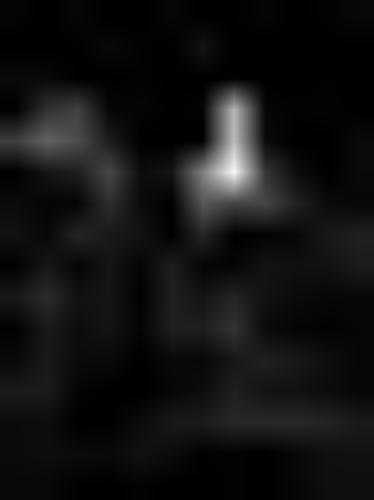}\\
	\vspace{0.05cm}
   \includegraphics[width=0.242\linewidth]{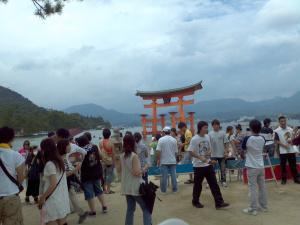}
   \includegraphics[width=0.242\linewidth]{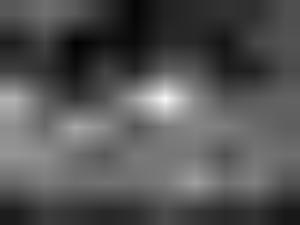}
   \includegraphics[width=0.242\linewidth]{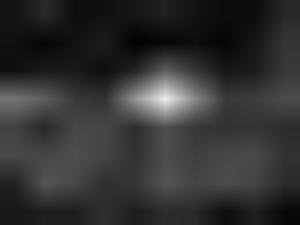}
   \includegraphics[width=0.242\linewidth]{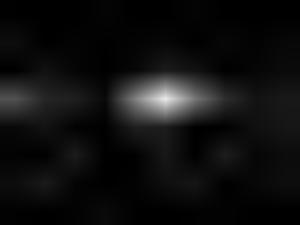}\\
	\vspace{0.05cm}
   \includegraphics[width=0.242\linewidth]{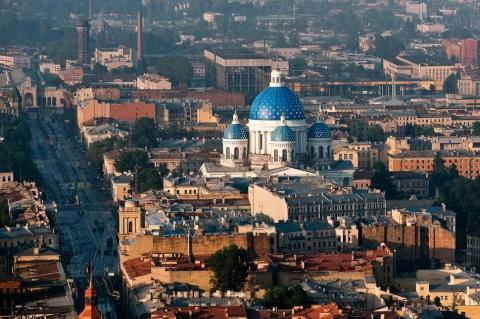}
   \includegraphics[width=0.242\linewidth]{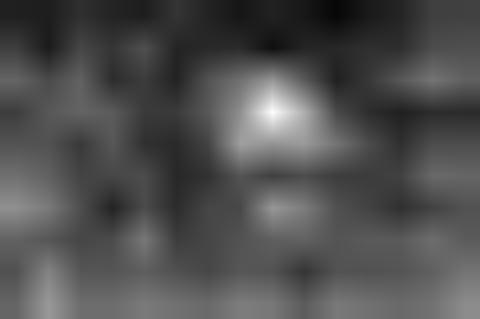}
   \includegraphics[width=0.242\linewidth]{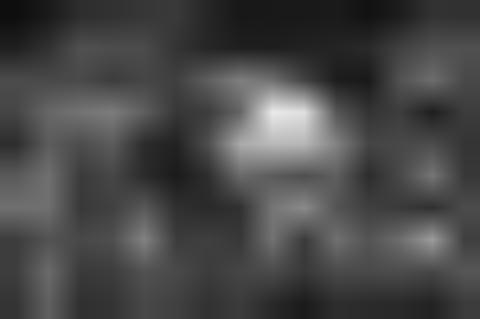}
   \includegraphics[width=0.242\linewidth]{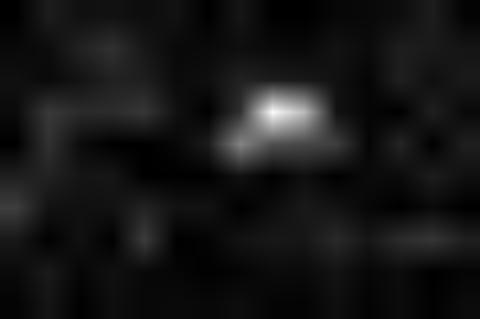}\\
	\vspace{-0.1cm}
\centerline{\footnotesize (a) \hspace{1.7cm} (b) \hspace{1.7cm} (c) \hspace{1.7cm} (d)}
	\vspace{-0.5cm}
\end{center}
	\caption{Comparison of keypoint selection methods. (a) Input image (b) $L_2$ norm scores using the pretrained model (DELF-noFT) (c) $L_2$ norm scores using fine-tuned descriptors (DELF+FT) (d) Attention-based scores (DELF+FT+ATT).  Our attention-based model effectively disregards clutter compared to other options.}
\label{fig:vis_att}
\vspace{-20pt}                                                                                                                                                                                                           
\end{figure}

\vspace{-3pt}
\paragraph{DELF vs. CONGAS}
%
The main advantage of DELF over CONGAS is its recall; it retrieves more relevant landmarks than CONGAS, which suggests that DELF descriptors are more discriminative.
We did not observe significant examples where CONGAS outperforms DELF.
\figref{fig:feat_matches} shows pairs of images from query and database, which are successfully matched by DELF but missed by CONGAS, where feature correspondences are presented by connecting the center of the receptive fields for matching features.
Since the receptive fields can be fairly large, some features seem to be localized in undiscriminative regions, \eg, ocean or sky.
However, in these cases, the features take into account more discriminative regions in the neighborhood.

\vspace{-7pt}
\paragraph{Analysis of keypoint detection methods}
\figref{fig:vis_att} visualizes three variations of keypoint detection, where the benefit of our attention model is clearly illustrated qualitatively while the $L_2$ norm of fine-tuned features is marginally different from the one without fine-tuning.

\subsection{Results in Existing Datasets}
\label{sub:results}
We demonstrate the performance of DELF in existing datasets such as Oxf5k, Par6k and their extensions, Oxf105k and Par106k, for completeness.
For this experiment, we simply obtain the score per image using the proposed method, and make a late fusion with the score from DIR by computing a weighted mean of two normalized scores, where the weight for DELF is set to 0.25.
The results are presented in \tabref{tab:100k_results}.
We present accuracy of existing methods in their original papers and our reproductions using public source codes, which are very close.
DELF improves accuracy nontrivially in the datasets when combined with DIR, although it does not show the best performance by itself.
This fact indicates that DELF has capability to encode complementary information that is not available in global feature descriptors.%
\begin{table}[t]
\footnotesize                                                                                                                                                                                                                
\centering     
\caption{Performance evaluation on existing datasets in mAP ($\%$).  All results of existing methods are based on our reproduction using public source codes.  We tested LIFT only on Oxf5k and Par6k due to its slow speed.  (* denotes the results from the original papers.)}
\vspace{-5pt}
\begin{tabular}{@{\hskip0.0pt}l@{\hskip2.5pt}|C{0.7cm}C{0.7cm}C{0.7cm}C{0.7cm}l@{\hskip0.0pt}}         
  \toprule                                                                                                                                                                                                                                                                      
Dataset & Oxf5k & Oxf105k & Par6k & Par106k \\     
  \midrule     
DIR~\cite{gordo2016deep} & 86.1 &  82.8   &    94.5     &     90.6   \\
DIR+QE~\cite{gordo2016deep} & 87.1 & 85.2 &   95.3    &   91.8   \\
siaMAC~\cite{radenovic2016cnn} & 77.1 & 69.5   & 83.9     & 76.3      \\
siaMAC+QE~\cite{radenovic2016cnn} & 81.7 & 76.6   & 86.2         &  79.8      \\
CONGAS~\cite{buddemeier2012systems} & 70.8 & 61.1 & 67.1 &  56.8  \\
LIFT~\cite{yi2016lift} & 54.0 & -- & 53.6 &  --  \\
\midrule
DIR+QE*~\cite{gordo2016deep} & 89.0 & 87.8 &   93.8     &   90.5   \\
siaMAC+QE*~\cite{radenovic2016cnn} & 82.9 & 77.9   & 85.6         &  78.3      \\
\midrule
DELF+FT+ATT (ours) & 83.8 & 82.6 & 85.0 & 81.7 \\
DELF+FT+ATT+DIR+QE (ours) & {\bf 90.0} & {\bf 88.5} & {\bf 95.7} & {\bf 92.8} \\
  \bottomrule            
\end{tabular}                                                                                                                                                                                                                                                                        
\label{tab:100k_results}    
\vspace{-10pt}                                                                                                                                                                                                           
\end{table}


\section{Conclusion}
We presented DELF, a new local feature descriptor that is designed specifically for large-scale image retrieval applications. DELF is learned with weak supervision, using image-level labels only, and is coupled with our new attention mechanism for semantic feature selection. 
In the proposed CNN-based model, one forward pass over the network is sufficient to obtain both keypoints and descriptors. 
To properly evaluate performance of large-scale image retrieval algorithm, we introduced Google-Landmarks dataset, which consists of more than 1M database images, 13K unique landmarks, and 100K query images. 
The evaluation in such a large-scale setting shows that DELF outperforms existing global and local descriptors by substantial margins. We also present results on existing datasets, and show that DELF achieves excellent performance when combined with global descriptors.

\paragraph{Acknowledgement}
This work was performed while the first and last authors were in Google Inc., CA.
It was partly supported by the ICT R\&D program of MSIP/IITP [2016-0-00563] and the NRF grant [NRF-2011-0031648] in Korea.

{\small
\bibliographystyle{ieee}
\bibliography{literature/delf}

\begin{thebibliography}{10}\itemsep=-1pt

\bibitem{aradhye2009video2text}
H.~Aradhye, G.~Toderici, and J.~Yagnik.
\newblock {Video2text: Learning to Annotate Video Content}.
\newblock In {\em Proc. IEEE International Conference on Data Mining
  Workshops}, 2009.

\bibitem{arandjelovic2015netvlad}
R.~Arandjelovi{\'c}, P.~Gronat, A.~Torii, T.~Pajdla, and J.~Sivic.
\newblock {NetVLAD: CNN Architecture for Weakly Supervised Place Recognition}.
\newblock In {\em Proc. CVPR}, 2016.

\bibitem{babenko2015iccv}
A.~Babenko and V.~Lempitsky.
\newblock {Aggregating Local Deep Features for Image Retrieval}.
\newblock In {\em Proc. ICCV}, 2015.

\bibitem{babenko2014neural}
A.~Babenko, A.~Slesarev, A.~Chigorin, and V.~Lempitsky.
\newblock {Neural Codes for Image Retrieval}.
\newblock In {\em Proc. ECCV}, 2014.

\bibitem{bay2008speeded}
H.~Bay, A.~Ess, T.~Tuytelaars, and L.~Van~Gool.
\newblock {Speeded-Up Robust Features (SURF)}.
\newblock {\em Computer Vision and Image Understanding}, 110(3):346--359, 2008.

\bibitem{beis1997shape}
J.~S. Beis and D.~G. Lowe.
\newblock {Shape Indexing Using Approximate Nearest-Neighbour Search in
  High-Dimensional Spaces}.
\newblock In {\em Proc. CVPR}, 1997.

\bibitem{bentley75multidimensional}
J.~L. Bentley.
\newblock {Multidimensional Binary Search Trees Used for Associative
  Searching}.
\newblock {\em Communications of the ACM}, 19(9), 1975.

\bibitem{buddemeier2012systems}
U.~Buddemeier and H.~Neven.
\newblock {Systems and Methods for Descriptor Vector Computation}, 2012.
\newblock US Patent 8,098,938.

\bibitem{dugas2001incorporating}
C.~Dugas, Y.~Bengio, C.~Nadeau, and R.~Garcia.
\newblock {Incorporating Second-Order Functional Knowledge for Better Option
  Pricing}.
\newblock In {\em Proc. NIPS}, 2001.

\bibitem{Fischler1981}
M.~Fischler and R.~Bolles.
\newblock {Random Sample Consensus: A Paradigm for Model Fitting with
  Applications to Image Analysis and Automated Cartography}.
\newblock {\em Communications of the ACM}, 24(6):381--395, 1981.

\bibitem{gordo2016deep}
A.~Gordo, J.~Almazan, J.~Revaud, and D.~Larlus.
\newblock {Deep Image Retrieval: Learning Global Representations for Image
  Search}.
\newblock In {\em Proc. ECCV}, 2016.

\bibitem{han2015matchnet}
X.~Han, T.~Leung, Y.~Jia, R.~Sukthankar, and A.~C. Berg.
\newblock {MatchNet: Unifying Feature and Metric Learning for Patch-Based
  Matching}.
\newblock In {\em Proc. CVPR}, 2015.

\bibitem{he2015deep}
K.~He, X.~Zhang, S.~Ren, and J.~Sun.
\newblock {Deep Residual Learning for Image Recognition}.
\newblock In {\em Proc. CVPR}, 2016.

\bibitem{hong2015decoupled}
S.~Hong, H.~Noh, and B.~Han.
\newblock {Decoupled Deep Neural Network for Semi-supervised Semantic
  Segmentation}.
\newblock In {\em Proc. NIPS}, 2015.

\bibitem{jegou2012negative}
H.~J{\'e}gou and O.~Chum.
\newblock {Negative Evidences and Co-Occurences in Image Retrieval: The Benefit
  of PCA and Whitening}.
\newblock In {\em Proc. ECCV}, 2012.

\bibitem{Jegou2008}
H.~J\'{e}gou, M.~Douze, and C.~Schmid.
\newblock {Hamming Embedding and Weak Geometric Consistency for Large Scale
  Image Search}.
\newblock In {\em Proc. ECCV}, 2008.

\bibitem{jegou2011product}
H.~Jegou, M.~Douze, and C.~Schmid.
\newblock {Product Quantization for Nearest Neighbor Search}.
\newblock {\em IEEE Transactions on Pattern Analysis and Machine Intelligence},
  33(1), 2011.

\bibitem{Jegou2010}
H.~J\'{e}gou, M.~Douze, C.~Schmidt, and P.~Perez.
\newblock {Aggregating Local Descriptors into a Compact Image Representation}.
\newblock In {\em Proc. CVPR}, 2010.

\bibitem{Jegou2012}
H.~J\'{e}gou, F.~Perronnin, M.~Douze, J.~Sanchez, P.~Perez, and C.~Schmid.
\newblock {Aggregating Local Image Descriptors into Compact Codes}.
\newblock {\em IEEE Transactions on Pattern Analysis and Machine Intelligence},
  34(9), 2012.

\bibitem{kalantidis2014locally}
Y.~Kalantidis and Y.~Avrithis.
\newblock {Locally Optimized Product Quantization for Approximate Nearest
  Neighbor Search}.
\newblock In {\em Proc. CVPR}, 2014.

\bibitem{kalantidis2015cross}
Y.~Kalantidis, C.~Mellina, and S.~Osindero.
\newblock {Cross-Dimensional Weighting for Aggregated Deep Convolutional
  Features}.
\newblock In {\em Proc. ECCV Workshops}, 2015.

\bibitem{Lowe2004}
D.~Lowe.
\newblock {Distinctive Image Features from Scale-Invariant Keypoints}.
\newblock {\em International Journal of Computer Vision}, 60(2), 2004.

\bibitem{neven2008image}
H.~Neven, G.~Rose, and W.~G. Macready.
\newblock {Image Recognition with an Adiabatic Quantum Computer I. Mapping to
  Quadratic Unconstrained Binary Optimization}.
\newblock {\em arXiv:0804.4457}, 2008.

\bibitem{yue2015exploiting}
J.~Y.-H. Ng, F.~Yang, and L.~S. Davis.
\newblock {Exploiting Local Features from Deep Networks for Image Retrieval}.
\newblock In {\em Proc. CVPR Workshops}, 2015.

\bibitem{Nister}
D.~Nist{\'e}r and H.~Stewenius.
\newblock {Scalable Recognition with a Vocabulary Tree}.
\newblock In {\em Proc. CVPR}, 2006.

\bibitem{perronnin2009}
F.~Perronnin, Y.~Liu, and J.-M. Renders.
\newblock {A Family of Contextual Measures of Similarity between Distributions
  with Application to Image Retrieval}.
\newblock In {\em Proc. CVPR}, 2009.

\bibitem{Philbin07}
J.~Philbin, O.~Chum, M.~Isard, J.~Sivic, and A.~Zisserman.
\newblock {Object Retrieval with Large Vocabularies and Fast Spatial Matching}.
\newblock In {\em Proc. CVPR}, 2007.

\bibitem{Philbin2008}
J.~Philbin, O.~Chum, M.~Isard, J.~Sivic, and A.~Zisserman.
\newblock {Lost in Quantization: Improving Particular Object Retrieval in Large
  Scale Image Databases}.
\newblock In {\em Proc. CVPR}, 2008.

\bibitem{radenovic2016cnn}
F.~Radenovi{\'c}, G.~Tolias, and O.~Chum.
\newblock {CNN Image Retrieval Learns from BoW: Unsupervised Fine-Tuning with
  Hard Examples}.
\newblock In {\em Proc. ECCV}, 2016.

\bibitem{ren2015faster}
S.~Ren, K.~He, R.~Girshick, and J.~Sun.
\newblock {Faster R-CNN: Towards Real-Time Object Detection with Region
  Proposal Networks}.
\newblock In {\em Proc. NIPS}, 2015.

\bibitem{russakovsky2015imagenet}
O.~Russakovsky, J.~Deng, H.~Su, J.~Krause, S.~Satheesh, S.~Ma, Z.~Huang,
  A.~Karpathy, A.~Khosla, M.~Bernstein, et~al.
\newblock {ImageNet Large Scale Visual Recognition Challenge}.
\newblock {\em International Journal of Computer Vision}, 115(3), 2015.

\bibitem{Simonyan15}
K.~Simonyan and A.~Zisserman.
\newblock {Very Deep Convolutional Networks for Large-Scale Image Recognition}.
\newblock In {\em Proc. ICLR}, 2015.

\bibitem{Sivic2003}
J.~Sivic and A.~Zisserman.
\newblock {Video Google: A Text Retrieval Approach to Object Matching in
  Videos}.
\newblock In {\em Proc. ICCV}, 2003.

\bibitem{tolias2015particular}
G.~Tolias, R.~Sicre, and H.~J{\'e}gou.
\newblock {Particular Object Retrieval with Integral Max-Pooling of CNN
  Activations}.
\newblock In {\em Proc. ICLR}, 2015.

\bibitem{torii13visual}
A.~Torii, J.~Sivic, T.~Pajdla, and M.~Okutomi.
\newblock {Visual Place Recognition with Repetitive Structures}.
\newblock In {\em Proc. CVPR}, 2013.

\bibitem{uricchio2015fisher}
T.~Uricchio, M.~Bertini, L.~Seidenari, and A.~Bimbo.
\newblock {Fisher Encoded Convolutional Bag-of-Windows for Efficient Image
  Retrieval and Social Image Tagging}.
\newblock In {\em Proc. ICCV Workshops}, 2015.

\bibitem{verdie2015tilde}
Y.~Verdie, K.~M. Yi, P.~Fua, and V.~Lepetit.
\newblock {TILDE: A Temporally Invariant Learned Detector}.
\newblock In {\em Proc. CVPR}, 2015.

\bibitem{xu2015show}
K.~Xu, J.~Ba, R.~Kiros, K.~Cho, A.~Courville, R.~Salakhudinov, R.~Zemel, and
  Y.~Bengio.
\newblock {Show, Attend and Tell: Neural Image Caption Generation with Visual
  Attention}.
\newblock In {\em Proc. ICML}, 2015.

\bibitem{yang2016stacked}
Z.~Yang, X.~He, J.~Gao, L.~Deng, and A.~Smola.
\newblock {Stacked Attention Networks for Image Question Answering}.
\newblock In {\em Proc. CVPR}, 2016.

\bibitem{yi2016lift}
K.~M. Yi, E.~Trulls, V.~Lepetit, and P.~Fua.
\newblock {LIFT: Learned Invariant Feature Transform}.
\newblock In {\em Proc. ECCV}, 2016.

\bibitem{moo2016learning}
K.~M. Yi, Y.~Verdie, P.~Fua, and V.~Lepetit.
\newblock {Learning to Assign Orientations to Feature Points}.
\newblock In {\em Proc. CVPR}, 2016.

\bibitem{zagoruyko2015learning}
S.~Zagoruyko and N.~Komodakis.
\newblock {Learning to Compare Image Patches via Convolutional Neural
  Networks}.
\newblock In {\em Proc. CVPR}, 2015.

\bibitem{zheng2016sift}
L.~Zheng, Y.~Yang, and Q.~Tian.
\newblock {SIFT Meets CNN: A Decade Survey of Instance Retrieval}.
\newblock {\em arXiv:1608.01807}, 2016.

\bibitem{zheng2009tour}
Y.-T. Zheng, M.~Zhao, Y.~Song, H.~Adam, U.~Buddemeier, A.~Bissacco, F.~Brucher,
  T.-S. Chua, and H.~Neven.
\newblock {Tour the World: Building a Web-Scale Landmark Recognition Engine}.
\newblock In {\em Proc. CVPR}, 2009.

\bibitem{zhou2016learning}
B.~Zhou, A.~Khosla, A.~Lapedriza, A.~Oliva, and A.~Torralba.
\newblock {Learning Deep Features for Discriminative Localization}.
\newblock In {\em CVPR}, 2016.

\end{thebibliography}
}

\end{document}